\ifdefined\XeTeXversion\else\ifdefined\pdfoutput \pdfoutput=1 \fi\fi
\documentclass{article}

\usepackage[preprint]{neurips_2026}

\usepackage[utf8]{inputenc}
\usepackage[T1]{fontenc}
\usepackage{hyperref}
\usepackage{url}
\usepackage{booktabs}
\usepackage{amsfonts}
\usepackage{amsmath}
\usepackage{amssymb}
\usepackage{nicefrac}
\usepackage{microtype}
\usepackage{graphicx}
\usepackage{multirow}
\usepackage{array}
\usepackage{xcolor}
\usepackage{enumitem}
\usepackage{listings}
\usepackage{needspace}
\usepackage{placeins}
\usepackage[font=small,labelfont=bf]{caption}

\lstdefinestyle{prompt}{%
  basicstyle=\scriptsize\ttfamily,
  breaklines=true,
  breakautoindent=false,
  breakindent=0pt,
  columns=fullflexible,
  keepspaces=true,
  frame=single,
  framesep=4pt,
  rulecolor=\color{black!35},
  xleftmargin=2pt,
  xrightmargin=2pt,
  aboveskip=6pt,
  belowskip=6pt,
  showstringspaces=false,
  upquote=true,
}

\graphicspath{{figures/}}

\definecolor{echogreen}{RGB}{16,80,64}

\newcolumntype{L}[1]{>{\raggedright\arraybackslash}p{#1}}

\newcommand{\envshot}[2]{%
  \begin{minipage}[t]{0.329\linewidth}%
    \centering
    \setlength{\fboxsep}{0pt}%
    \setlength{\fboxrule}{0.3pt}%
    \fbox{\includegraphics[width=\dimexpr\linewidth-2\fboxrule\relax]{#1}}\\[1.5pt]
    {\scriptsize #2}%
  \end{minipage}%
}

\renewcommand{\acksection}{\section*{Acknowledgements}}

\makeatletter
\renewcommand{\@notice}{\enlargethispage{2\baselineskip}}
\makeatother

\newcommand{\pibase}{\pi_{\mathrm{base}}}
\newcommand{\pisft}{\pi_{\mathrm{SFT}}}
\newcommand{\pirl}{\pi_{\mathrm{RL}}}
\newcommand{\pishallow}{\pi_{\mathrm{shallow}}}
\newcommand{\pideep}{\pi_{\mathrm{deep}}}
\newcommand{\echo}[1]{\textsc{#1}}

\title{Echoverse: Deep, Evolving Environments for\\Training Computer-Use Agents at Scale}

\author{%
  \bfseries
  Yash Pandya \quad
  Sahil Gupta\thanks{Work done while at Microsoft Research.} \quad
  Sarthak Harne \quad
  Archana Yadav\footnotemark[1] \quad
  Kavyansh Chourasia \\
  \bfseries
  Hussein Mozannar \quad
  Vibhav Vineet \quad
  Sara Abdali \quad
  Corby Rosset \\
  \bfseries
  Yash Lara \quad
  Ahmed Awadallah \quad
  Ece Kamar \quad
  Akshay Nambi\thanks{Corresponding author: \texttt{akshayn@microsoft.com}} \\[2pt]
  Microsoft Research \\
}

\begin{document}

\maketitle

\begin{abstract}
Computer-use agents learn from what their actions change, so training one needs
applications it can act on, break and reset. The ones that matter most are
login-gated and stateful, so synthetic environments stand in for them. Recent
pipelines generate such environments in bulk, moving the bottleneck from how
many exist to what is inside each one. The returns come from three properties:
how much behavioural depth an environment carries, whether it targets the
interaction an agent actually fails, and whether it improves alongside the
model. We present Echoverse, which compiles specifications into stateful
applications whose tasks are graded against the application's own database, and
a co-evolution loop that reads every graded rollout twice: as repairs to the
environment, its tasks and its verifier, and as training signal. Trained on
twelve such environments, a 9B model improves from $36.5\%$ to $67.1\%$ across
fourteen evaluation splits, within fourteen points of the much larger frontier
model that taught it. Taking the three properties in turn: on the same domains,
shallow environments push live-site accuracy below the base model
($80.0 \to 75.0$) while deep ones raise it ($80.0 \to 85.0$ and
$48.0 \to 65.0$); drilling one interface control across many renderings
transfers to held-out widget families and to the open web; and repairing a
single environment lifts the model trained on it from $16.2\%$ to $38.5\%$. The
same worlds serve as reinforcement-learning environments: a reward combining the
grounded verifier with a dense per-step judge raises held-out score from
$58.8\%$ to $68.0\%$. We release four environments as a benchmark, with their
applications, seed data and graders. Code: \url{https://aka.ms/echoverse}.
\end{abstract}

\section{Introduction}
\label{sec:intro}

A computer-use agent learns the results of its actions only where those actions
have consequences. A click changes saved state, a message reaches a real
person, or a page that refuses to move tells the agent that its last move did
nothing. A screenshot shows what an interface looks like; only a running
application shows what an action caused.

The consequences worth learning from are stateful, and most of them sit behind
a login: email and chat, banking, health records, the internal consoles for
cloud and machine learning. None of these can be trained against directly.
Every attempt writes to a real account, there is no reset between tries, and the
true state stays hidden behind the screen. We instead rebuild the application as
a synthetic environment whose database we own, so that state changes for real
but is safe to break, quick to reset, and gradable from the data rather than
from a screenshot.

By a \emph{world} we mean three things bound together: an \textbf{environment}
(the application, its state, and the actions that change it), the \textbf{tasks}
that set goals in it, and a \textbf{verifier} that grades the outcome against
ground truth. Recent pipelines produce such worlds in bulk, yielding hundreds of
environments and thousands of checkable tasks
\citep{webarenainfinity,infiniteweb,gymanything,cuagym}, and this work builds on
that progress. Once worlds are plentiful, the bottleneck moves inside them, into
whether each one holds together under real use, which is a question of depth
rather than count: whether state stays coherent across users and screens,
whether workflows keep their dependencies, whether a weak skill recurs in enough
forms to generalise, and whether success is judged by outcome or by appearance.

\paragraph{Thesis.} We argue that the gains come less from adding worlds than
from a loop that keeps improving the ones already built. Echoverse treats
building the environment and training the model as one process rather than two
stages: run a model in a world, find where it fails, make the environment, its
tasks and its verifiers more faithful or more demanding at that point, train on
the sharper signal, and repeat. Ordinary fine-tuning improves only the model.
Here the same graded run that measures the model also improves the world that
judged it, so a static benchmark saturates while the loop compounds.

\begin{figure}[tbp]
  \centering
  \includegraphics[width=\linewidth]{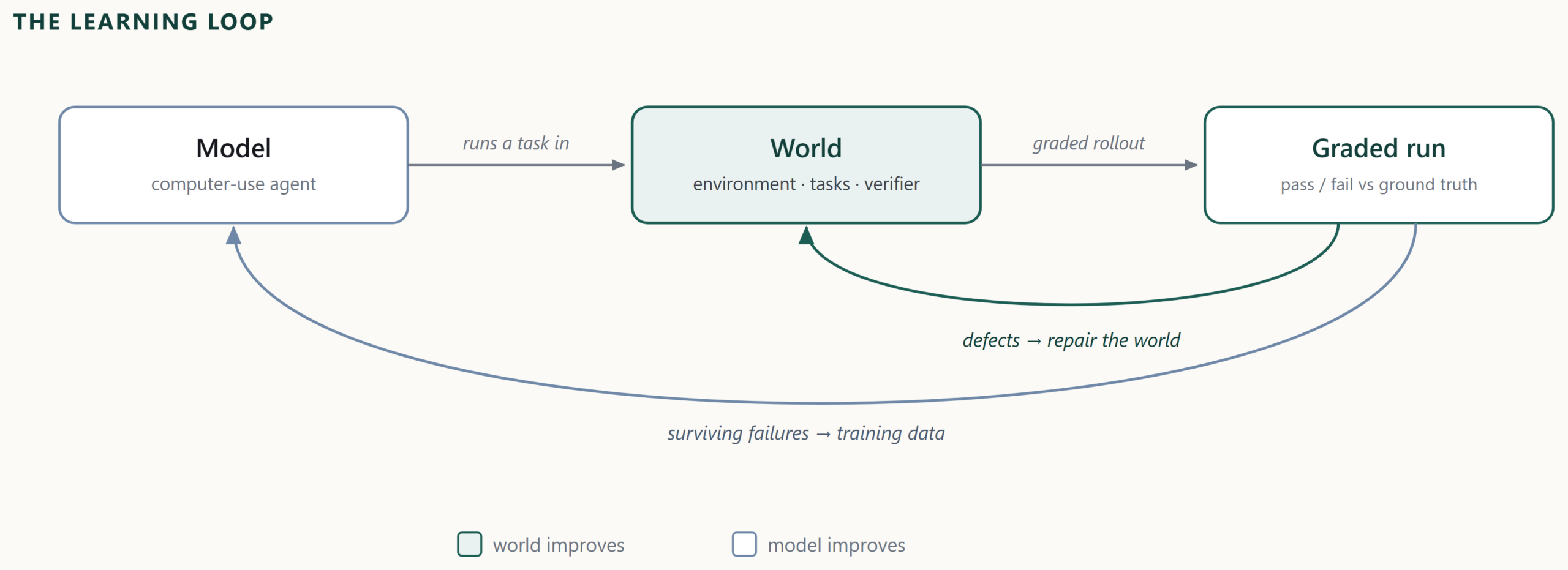}
  \caption{The learning loop. Every graded run is read twice. Failures that
  survive triage become model training data; defects in the environment, its
  tasks or its verifier become repairs. The same graded run that measures the
  model also sharpens the world.}
  \label{fig:loop}
\end{figure}

Three levers keep the loop productive, none of them raw environment count.
\textbf{Depth}: worlds built backwards from the workflows they are meant to
teach, so that every one of those workflows can actually be carried out in them,
including in closed and proprietary domains. \textbf{Capability targeting}:
narrow worlds built around the exact interaction a model keeps failing.
\textbf{Co-evolution}: improving the environment, its tasks and its verifiers
on every graded run, not just the model. Figure~\ref{fig:loop} shows how the two
readings of a single graded run feed each other.

\paragraph{Reinforcement learning changes what is at stake.} For supervised
training, the argument above is one about data quality: a better world yields
better demonstrations. Under reinforcement learning it becomes a precondition.
Imitation carries a ceiling, because a clean demonstration never shows how to
recover from a mistake or when to stop, and those are the failures that break
agents in the wild. RL removes that ceiling but demands a volume of graded,
resettable episodes the live web cannot supply: it will not reset, it throttles
automated traffic far below the required rate, and it exposes no ground truth,
so reward must come from a second model judging a screenshot. Every property
that makes an Echoverse world good supervised data is also a prerequisite for
RL, so the worlds serve as reinforcement-learning environments without
modification. Sec.~\ref{sec:rl} develops the argument and reports a run over
five worlds that lifts the held-out judged score from $58.8\%$ to $68.0\%$. That
run uses a composite reward, grounded at the trajectory level and model-judged
per step, for reasons we set out in Sec.~\ref{sec:rl-method}.

\Needspace*{6\baselineskip}
\paragraph{Contributions.}
\begin{enumerate}[leftmargin=1.6em,itemsep=2pt,topsep=2pt]
\item \textbf{An operational definition of depth} (Sec.~\ref{sec:setting}):
depth as completeness with respect to a target workflow set rather than as
feature count, decomposed into five properties and discharged at build time
against machine-checkable claims, so that ``deep'' is something a world is shown
to be rather than something claimed about it.
\item \textbf{The Echoverse factory} (Sec.~\ref{sec:factory}): a two-phase
pipeline that compiles seeds into an application verified against
machine-checkable claims, then grows a task corpus regrounded on live data and
proven solvable through the real user interface, every task carrying a
database-grounded verifier.
\item \textbf{The co-evolution loop} (Sec.~\ref{sec:loop}): one graded rollout,
read as two orthogonal signals, under an ordering in which the world is repaired
before the model's failures are trusted, and under a discipline that never
weakens a goal to raise a pass rate.
\item \textbf{Evidence for three levers} (Sec.~\ref{sec:experiments}),
including the negative result that shallow worlds are worse than not training at
all, and the finding that added environments keep helping after added
trajectories have stopped.
\item \textbf{Echoverse as a reinforcement-learning environment}
(Sec.~\ref{sec:rl}): the same worlds, unmodified, meet the reset, throughput and
reward requirements of RL, and a run over five of them trains a policy beyond
its supervised initialisation.
\item \textbf{A public benchmark release} (Sec.~\ref{sec:release}): four worlds,
the deep domains \echo{EchoStay} and \echo{EchoForge} and the two capability
worlds with their held-out splits, each shipping as a runnable application with
its seed database and, for every released task, the grounded verifier that
scores it. We release evaluation tasks only, so the benchmark measures agents
rather than doubling as training data for them. Environment code and graded
tasks are at \url{https://aka.ms/echoverse}.
\end{enumerate}

\section{Related work}
\label{sec:related}

\paragraph{Live-web benchmarks.} WebVoyager \citep{webvoyager2024},
Mind2Web \citep{mind2web2023} and Online-Mind2Web \citep{onlinemind2web2025}
evaluate agents against real public websites and remain the best available
measure of open-web competence. Their coverage is bounded by what an
unauthenticated visitor can do. No benchmark can hand out real credentials or
let an agent write to a stranger's account, so the tasks come almost entirely
from the logged-out surface of the web and are dominated by information finding:
locate a page and read a value off it. The workflows that motivate computer-use
agents sit on the other side of the login, where an action commits a booking or
moves money and success is a change to stored state rather than a string the
agent reports. Their verdicts are also model-generated. With no access to a
site's backend, both judge success from screenshots and text with a multimodal
LLM \citep{judging2023}, so a score is an opinion about appearance rather than a
fact about application state, and the characteristic error is accepting an
agent's confident claim to have completed an action that never landed. That is
tolerable noise for a leaderboard read in aggregate, but it rules the signal out
as a training reward (Sec.~\ref{sec:rl-why}). These benchmarks also make poor
training grounds because they do
not hold still: pages are redesigned, listings and dates roll forward, and hosts
throttle or block automated traffic. An occasional evaluation can absorb that
drift; training cannot, since it runs the same task thousands of times and needs
the same world each time. We use these benchmarks only as held-out transfer
measurements (Sec.~\ref{sec:exp-liveweb}).

\paragraph{Synthetic environment suites.} WebArena \citep{webarena2024},
VisualWebArena \citep{visualwebarena2024}, WorkArena \citep{workarena2024},
OSWorld \citep{osworld2024} and AndroidWorld \citep{androidworld2025}
established hand-built, resettable environments with programmatic checks, and
BrowserGym \citep{browsergym2024} consolidated several of them behind one
harness. A more recent line automates their construction
\citep{webarenainfinity, infiniteweb, gymanything, cuagym}, and
UltraCUA \citep{ultracua} trains a foundation model over such data. These
pipelines scale environment count. Our contribution is complementary: we hold
count fixed and vary interior quality, and we show that below a depth threshold
additional environments contribute noise rather than signal.

\paragraph{What makes a training environment good.} The closest published
statement of requirements to ours comes from practice rather than from a
benchmark paper. \citet{amazoncuarl} list five properties a web ``gym'' needs in
order to drive learning: realism, explorability, data diversity and hydration,
correct verifiers, and infrastructure stability. Three of those correspond
directly to properties in Table~\ref{tab:depth}, and the two lists differ in
emphasis in a way worth naming. Theirs is written from the perspective of
keeping a large RL system running, so it includes infrastructure stability,
which we treat as engineering rather than as a property of a world. Ours is
written from the perspective of what a single episode teaches, so it includes
coherent cross-actor state and workflow dependency, which are exactly the
properties a shallow but visually convincing clone can fail while still passing
inspection. We share their conclusion that task design dominates task quantity;
Sec.~\ref{sec:exp-scaling} gives a measured version of it, in which environment
diversity continues to help while trajectory volume saturates.

\paragraph{Verification.} Tool-agent benchmarks such as
$\tau$-bench \citep{taubench2024} grade against database state rather than text
overlap, and we extend that tradition: our verdict of record is a database
round-trip, so an answer copied from the public web is rejected because it does
not reproduce. This matters more for training than for evaluation. A tolerant
LLM judge \citep{judging2023} used as a training filter teaches the policy the
judge's blind spots, and under RL the same failure is sharper still, since a
verifier that rewards work the agent did not do is a reward the optimiser will
find and exploit \citep{amazoncuarl}. A model still makes the final comparison
in our loop, since equivalence is not string identity, but it compares an
observed outcome against a reference read out of the database rather than
assessing a trajectory (Sec.~\ref{sec:grading}), which is a much narrower
question than the one such judges usually fail.
Sec.~\ref{sec:loop} treats the verifier as
an object of repair for this reason, alongside the environment and the task it
scores. Where ground truth is not
available, careful judge design remains the alternative;
\citet{universalverifier} show that separating process and outcome criteria and
attending to the whole trajectory drives judge false-positive rates close to
zero. Our two-term reward in Sec.~\ref{sec:rl-method} follows the same
process-plus-outcome split, with the outcome term grounded rather than judged.

\paragraph{Relation to our own prior work.} Fara-7B \citep{fara7b} and
Fara-1.5 \citep{fara15} are our earlier computer-use agents, trained on data
from a generation pipeline that draws on both live websites and synthetic
environments; the synthetic component of that pipeline used an early version of
the worlds described here. Those papers report end-to-end models built from a
mixed pipeline. This paper takes the environments themselves as the object of
study and asks which of their properties make a world worth training on, which
is why the models reported here are trained on synthetic data alone and are not
comparable to the Fara releases.

\paragraph{Curriculum and task curation.} Which tasks to spend RL budget on is
its own design problem, because tasks the policy already solves produce no
gradient and tasks it never solves produce noise. \citet{amazoncuarl} track
per-task success rates and concentrate sampling in a mid-range band of roughly
30 to 70 percent. We apply the same idea as a one-off filter rather than as a
running curriculum: we sample the starting policy four times per task and keep
only the tasks it solves once, twice or three times, discarding both the ones it
never solves and the ones it always solves
(Sec.~\ref{sec:rl-details}).

\paragraph{Automatic curriculum and environment design.} Co-evolution is
adjacent to open-ended and unsupervised environment design
\citep{poet2019,paired2020,accel2022,curriculumsurvey2020} and to
self-play \citep{alphazero2018}. The difference is what is evolved. Those
methods search over a parameterised task distribution inside a fixed, correct
simulator. We co-evolve the implementation of the environment, its task corpus
and its verifier, because in generated software all three are defective at the
start and a defect in any one silently poisons measurement of the others.

\paragraph{Distillation and reinforcement learning.} Our supervised stage is
verifier-filtered rejection-sampled distillation, following
STaR \citep{star2022} and rejection fine-tuning \citep{rft2023}, and inherits
the corresponding teacher ceiling. Our RL stage uses a group-relative policy
gradient \citep{deepseekmath2024,deepseekr1} with the unnormalised advantage of
\citet{drgrpo2025}, which suits sparse, verifier-supplied reward over long
browser trajectories. The practical obstacles in this regime, namely the
train-inference gap, credit assignment over hundreds of steps, and rollout
throughput, are documented in \citet{amazoncuarl}. What distinguishes our
setting is not the algorithm but the environment requirement
(Sec.~\ref{sec:rl-why}).

\section{Problem setting: environments, tasks and grounded grading}
\label{sec:setting}

\subsection{Environments, tasks and worlds}

An \emph{environment} is a tuple
$E = (\mathcal{S}, \mathcal{A}, T, \mathcal{O}, D_0)$.
The state space $\mathcal{S}$ is the application's database, not its rendering;
$\mathcal{A}$ is a browser action space; $T$ is the application's own transition
function, realised by its backend; $\mathcal{O}$ maps state to what the agent
sees; and $D_0$ is the seeded initial database. Taking the state to be the
database rather than the pixels is what makes reward groundable and reset exact,
and the rest of the paper follows from it.

A \emph{task} is a tuple $\tau = (g, r, \kappa)$, where $g$ is a
natural-language goal,
$\kappa \in \{\textsc{read}, \textsc{write}, \textsc{read\_write}\}$ is the task
kind, and $r$ is the reference the outcome is checked against: a \emph{reference
answer} for a \textsc{read} task, a \emph{reference state change} for a
\textsc{write} task, and both for a \textsc{read\_write} task. References are
minted from $D_0$ at generation time by executing SQL against the live database,
so a task is true by construction rather than by assertion.

A \emph{world} $W = (E, \mathcal{T}, V)$ is an environment together with a task
corpus $\mathcal{T}$ and the verifier $V$ that grades outcomes in it, as
introduced in Sec.~\ref{sec:intro}. The distinction matters throughout: the
environment is the application, while the world is the trainable unit, and the
co-evolution loop of Sec.~\ref{sec:loop} repairs all three components.

\subsection{Grounded grading}
\label{sec:grading}

Let $D_T$ denote the database after the agent terminates. Grading is defined per
task kind:
\begin{itemize}[leftmargin=1.4em,itemsep=1pt,topsep=2pt]
\item \textsc{write} is graded against the reference state change: it must be
      present in the \texttt{sqldiff} between the pristine per-task copy $D_0$
      and $D_T$, which establishes both that the write landed and that it is the
      write the task asked for, and stops a task passing on state that was
      already there.
\item \textsc{read} is graded against the reference answer, which was itself
      read back from the database at generation time. The agent's
      response must match it under semantic equivalence (\$288 for \$287.62
      passes).
\item \textsc{read\_write} carries both references and scores the minimum of the
      two.
\end{itemize}

\paragraph{How the comparison is made.} Neither check is a string match. An
answer of \$288 should count against a reference of \$287.62, and a booking
written with the right listing, dates and guest count should count whether or
not incidental fields differ. We therefore make the final comparison with an LLM
under fixed guidelines, reproduced in full in Appendix~\ref{app:prompts}, and it
is worth being precise about what that judge does
and does not see. Its input is two items: the agent's reported answer, or the
\texttt{sqldiff} between $D_0$ and $D_T$ for a write, together with the
corresponding reference. It
does not receive the trajectory, the screenshots or the agent's own account of
what it did, and it is never asked whether the agent behaved sensibly, only
whether the observed outcome matches the reference it is given. Changes beyond
those the reference requires are tolerated, so an agent that also stars an email
while marking it read is not penalised for it. The call is
therefore small and cheap enough to run on every rollout of a training run.

Reliability follows from that narrowness rather than from the judge's strength.
Deciding whether two amounts agree, or whether an observed row diff carries the
same content as a reference one, is a closed comparison against a value that was
itself read out of the database. That is a far easier question than the one a
live-web judge must answer, which is whether a fifty-step episode succeeded
given only screenshots and the agent's own summary, with no ground truth to
compare against. The grounding lives in where the reference comes from and in
how little the judge is asked to decide.

Grading is therefore hard to game, grounded rather than labelled, and uniform
across an \echo{EchoStay} booking, an \echo{EchoForge} issue and an
\echo{EchoBank} transfer. It eliminates the two dominant false-positive modes we
see in practice: the agent claiming a success it did not achieve, and a tolerant
judge accepting an answer retrieved from the public web.

\subsection{Depth, defined operationally}
\label{sec:depth}

An environment can be stable, plausibly styled and still be useless to train on.
What separates a world worth training on from one that is not is depth, and
depth is not an absolute quantity. A world is
deep when its environment supports every workflow its task corpus demands, end
to end through the interface, and each leaves an effect in state precise enough
to check.

Completeness is therefore relative rather than exhaustive. We do not try to
reproduce every feature of a banking product, only every action its transfers,
bill payments and disputes actually need, with the balances, permissions and
histories those actions touch. An environment missing a feature no target
workflow exercises is not shallow; an environment whose booking flow cannot set
a guest count is shallow however many pages it renders.

This is what makes the definition operational. The factory takes the workflows
as an input rather than discovering them afterwards: scenarios, seed data,
diagnosed capability gaps and the intended tasks all enter at the start
(Sec.~\ref{sec:factory}), are compiled into machine-checkable claims about the
routes and state they require, and the database, backend and frontend are
repaired until those claims pass and the tasks are completable by an agent
driving the real interface. ``Deep'' is then a statement about what a world has
been shown to support rather than a property asserted for it.

\begin{table}[htbp]
\centering
\small
\caption{What supporting a target workflow set requires. The first four are
structural obligations on the world; the fifth decides which workflows are worth
imposing them for.}
\label{tab:depth}
\begin{tabular}{@{}p{3.6cm}p{9.5cm}@{}}
\toprule
\textbf{Property} & \textbf{What it requires} \\
\midrule
Behavioural fidelity & Controls, permissions and errors follow the product's logic \\
Coherent state & A sent message appears for its recipient; a cancelled meeting clears both calendars \\
Workflow depth & An early choice constrains what happens later \\
Authoritative verification & Success is a property of application state, not of pixels \\
Domain value & The workflow is worth improving \\
\bottomrule
\end{tabular}
\end{table}

Table~\ref{tab:depth} decomposes that obligation. A workflow cannot
be driven if controls, permissions and errors do not follow the product's logic;
it cannot span actors if a sent message never reaches its recipient; it is not a
workflow at all if an early choice constrains nothing later; and it cannot be
graded if success is visible only in pixels. The fifth property is a choice
rather than a requirement, and decides where the other four are worth paying
for.

In the systems that matter most, the difficulty lives in permissions, shared
state and audit histories, which is exactly the structure a shallow clone skips.
Above this bar, more environments add variety; below it, they add noise, and
Sec.~\ref{sec:exp-depth} measures that difference directly. Depth also requires
the world to be fixed in time and data: the calendar does not move, the seed
data does not churn, and a task means the same thing on the thousandth rollout
as on the first. We trade a little surface realism for that control, and in
Sec.~\ref{sec:rl} the trade stops being a convenience and becomes the
precondition for training at all.

\section{Echoverse: the factory and the co-evolution loop}
\label{sec:factory}

\begin{figure}[tbp]
  \centering
  \includegraphics[width=\linewidth]{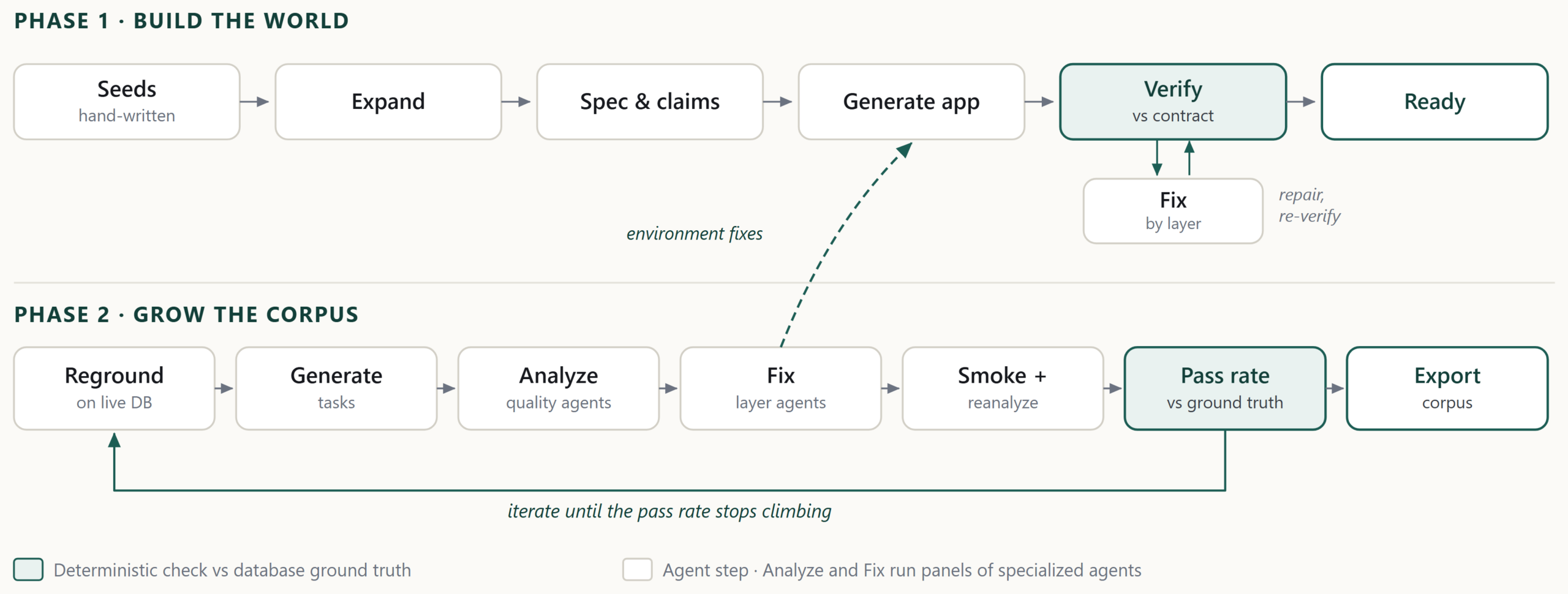}
  \caption{The environment factory. Phase 1 expands a handful of seeds into an
  application, then repairs the database, backend and frontend until it passes
  machine-checkable claims. Phase 2 regrounds tasks on live data, runs a panel
  of verifier, triage and layer-specific fixer agents, and re-scores against
  database ground truth until the corpus clears its verification bar. Many of
  those fixes land in the environment itself (dashed arrow).}
  \label{fig:pipeline}
\end{figure}

Echoverse is a single pipeline that produces two distinct types of outputs:
full-domain worlds that preserve the depth and structure of realistic workflows,
and capability worlds that isolate and vary a specific diagnosed interaction or
capability. Both rely on our owning the database underneath, so that success is
a property of the application's own state rather than of a model's reading of a
screenshot. Figure~\ref{fig:pipeline} shows the pipeline end to end, organised
into two phases.

In Phase~1 we build and validate the world itself. The process begins with
hand-written seeds describing representative scenarios, tasks, capabilities and
features. These seeds are expanded into a structured specification and a set of
explicit, verifiable claims, which are then used to generate the complete
application, including its frontend, backend and database. Generation is
followed by an iterative verification-and-repair cycle: claims are checked
against the generated application, failures are localised and fixed at the
appropriate layer, and the application is re-verified, until the specification
is satisfied and a validated environment remains.

In Phase~2 we grow a grounded task corpus on top of that environment. The
original seeds are regrounded against the live database to generate tasks whose
expected outcomes can be determined directly from application state. Each task,
together with the corresponding state of the environment, is then evaluated for
quality and solvability. Failures are diagnosed and repaired, and the tasks are
re-executed and re-scored against database-derived ground truth, until enough
tasks clear every verification criterion for the corpus to be exported.

The two phases run the same cycle, and are better read as one loop than as two
stages. Sec.~\ref{sec:loop} makes that loop the object of study, because it does
not stop when a world ships: the failures of the model being trained on a world
are themselves evidence about the world.

\subsection{An agent workforce}
\label{sec:agents}

Every stage in Figure~\ref{fig:pipeline} is carried out by an agent rather than
by a fixed script. The pipeline is implemented as a set of GitHub Copilot SDK
agents, each given a role, the repository of the world under construction, and
the tools an engineer would use in that role: a terminal for running the stack,
its migrations and its tests; Playwright \citep{playwright} for driving and
inspecting the running interface as a user meets it; and database tools exposed
over the Model Context Protocol \citep{mcp} for querying and editing the
application's own state. The roles divide into four kinds.

\begin{itemize}[leftmargin=1.4em,itemsep=1pt,topsep=2pt]
\item \textbf{Builders} produce the artefacts: schema and seed data, backend
      routes, the frontend that exercises them, and the task corpus.
\item \textbf{Verifiers} test what the builders produced, against the
      specification in Phase~1 and against quality and solvability in Phase~2.
\item \textbf{Triagers} read the verifiers' findings across the whole world at
      once, group failures that appear to share a cause, and turn them into a
      smaller set of consolidated work items, each attributed to the layer that
      must change.
\item \textbf{Fixers} take those items, debug against the running system, and
      re-check the repair, rolling it back if it regresses something else.
\end{itemize}

Two features of this arrangement do the work. Verification is separated from
construction, so no agent is the sole judge of its own output. And triage sits
between verification and repair, so fixes are planned over the whole set of
observed failures rather than one task at a time, which is what lets a single
repair clear many tasks (Sec.~\ref{sec:loop-repair}). Human supervision and QA
sit over the whole process as an additional layer of oversight on agent-driven
generation, verification and repair.

\subsection{Phase 1: building the environment}
\label{sec:phase1}

The pipeline expands a handful of seed scenarios into a specification, then
compiles it into machine-checkable claims about routes, state and behaviour.
Only then is the application generated: a FastAPI and SQLite backend under a
React interface. Verifier agents exercise every claim against the running
environment, driving the interface and reading the database behind it, and
fixers repair the database, backend or frontend until the claim holds. We
advance a world when at least $95\%$ of its claims pass; the remainder is
written into a \emph{readiness record} that separates hard blockers from
advisory risks, and an environment with an open blocker does not advance at all.
This is the first half of the
mechanism behind Sec.~\ref{sec:depth}: the claims are derived from the workflows
the world is meant to carry, so passing them is evidence that those workflows
are supported rather than merely rendered. Phase 2 supplies the second half by
requiring that the tasks themselves can be completed through the interface.

\subsection{Phase 2: growing the corpus}
\label{sec:phase2}

An environment that builds cleanly is still not training data. From the seeds, we create
tasks grounded on the live database, drawing goals from entities that actually exist, then
send every goal through a panel of verifiers. These
check whether its entities are real, whether the goal is plausible, whether its
difficulty matches the work, and, most consequentially, whether the goal is
feasible at all.

That last check is white-box, and deliberately so. The verifier agent drives the
running interface with Playwright, but it also reads the application's source
and queries its database directly, so it can separate a goal no interface can
satisfy from one that is merely hard. Feasibility is therefore a property of the
application, established by a factory agent, rather than a record of what some
particular policy happened to solve: no model that we later train or evaluate
takes part in deciding which tasks exist. This matters for how the evaluation
splits should be read, since a corpus filtered by a frontier model's success
would be a corpus shaped to that model's strengths.

Every failure becomes an issue tagged by the layer that must change: database,
backend, frontend, task text, or verifier. Triagers consolidate the issues that
share a cause, and layer-specific fixers apply the repair, re-check it, and roll
it back if it regresses. The loop re-scores against database ground truth until
at least $95\%$ of the surviving corpus passes verification, and each exported
task carries the exact check that grades it.

Fixing or improving the environment and growing the corpus are one loop rather than two
stages, and together they produce the world. In our experience most defects
belong to the environment rather than to the task text, so we re-version it on
every iteration. Harder tasks expose gaps in the environment, and a sturdier
environment can carry harder tasks.

\subsection{Seeding}

Where a rich public dataset exists we build on it. \echo{EchoStay} is seeded
from InsideAirbnb \citep{insideairbnb}, a public dataset compiled from publicly
visible short-term-rental listings, so its listings, hosts, reviews and
amenities are real rather than invented, and \echo{EchoForum} sits on a public
forum corpus of 2.55 million comments. Where none exists, as with mail,
calendar, banking and health records, a seeding pipeline generates the state
under strict constraints: dense and internally consistent, not a handful of
placeholder rows. Table~\ref{tab:suite-detail} gives per-world statistics.

\subsection{Determinism, isolation and reset}
\label{sec:reset}

Every rollout runs against an isolated per-task copy of the database, so no
task can contaminate another and the environment side of any run can be
reproduced exactly. Whenever seed data changes, the grounding database is
regenerated, re-validated and re-published, so corpus and shipped artefact never
drift. Sec.~\ref{sec:rl} reuses these reproducibility properties unchanged as the
reset primitive of a reinforcement-learning environment.

\subsection{One rollout, two signals}
\label{sec:loop}

The score an agent earns is never a measure of the model alone. It reflects a
coupled stack comprising the agent, the environment, the task, and the verifier.
A zero can therefore have many causes: the agent may have failed, but the
environment may also be broken, the requested state may be impossible to reach,
or the verifier may be checking the wrong condition. Treating every failure as
an agent capability gap turns defects in the synthetic world into erroneous
supervision. That limits what the agent can learn, and risks teaching it to
avoid otherwise valid features and interaction patterns simply because they were
broken in its training environments. We therefore treat every
graded rollout as a test of the entire stack. Failures attributable to the
environment, task, or verifier identify defects that should be repaired rather
than learned from. What remains, failures on tasks that have been established
to be valid, solvable, and correctly verified, is a cleaner signal
of the agent's actual capability gaps and therefore of the hard cases it should
train on next.

Attribution needs no separate apparatus. Failing rollouts go back through the
same verifier and triage agents the factory already runs, which ask of a rollout
what they ask of a candidate task: whether the goal is reachable through the
interface, whether the state it names exists, and whether the check that graded
it was the right one. The distinction we act on is coarse, between failures that
implicate the world and failures that implicate the model, and where the reading
is unclear we prefer to repair, since a task withheld from training costs less
than a task that teaches the wrong lesson.

\subsection{Repair before training, and why one fix clears many tasks}
\label{sec:loop-repair}

Repairing the world before training is what makes the resulting model signal
trustworthy. A failure caused by a broken environment is noise, not a useful
learning target. The principle is therefore not simply that generated
benchmarks contain bugs, but that these bugs must be identified and repaired
before agent failures are used as supervision. This ordering also imposes an
important discipline on the repair loop: goals are never weakened merely to
increase pass rates. Fixers may change the environment, the task text or the
verifier, but a repair that leaves a task asking for less than it did is treated
as a regression rather than a fix, and the bar a corpus must clear is the one
stated in Sec.~\ref{sec:phase2} rather than one relaxed to fit what the model
can already do. If a task is valid, solvable, and correctly verified, then a
model failure is precisely the signal we want to preserve.

Repair also has a multiplicative effect because failures that appear independent
at the task level often share a common underlying cause. Triage groups failures
by that cause and prioritises fixes according to how many tasks they affect.
This makes central environment defects particularly valuable repair targets:
a single broken control or interaction can block a large share of a corpus, and
restoring it restores all of those tasks at once. Sec.~\ref{sec:exp-coevo}
reports several such rounds, and what one of them does to the model trained on
the repaired world.

Many of these defects are easy to miss during ordinary human QA because they are
specific to how automated agents interact with the application. A human user,
for example, may set a guest count without giving the control a second thought.
If an agent cannot reliably operate that same control, however, it silently
leaves the default value unchanged, causing every booking task that depends on
guest count to fail. Agent-driven execution therefore exposes a class of
environment defects that may look innocuous to humans but systematically distort
the training signal for agents.

\subsection{From tasks to trajectories}
\label{sec:sft}

Only once a world has been through that loop do its tasks become training data,
and they do so through a single process. A frontier model
(GPT-5.4) solves each task in the live environment, the grounded verifier of
Sec.~\ref{sec:grading} keeps only the trajectories that pass, and those become
the supervised fine-tuning corpus behind every experiment in
Sec.~\ref{sec:experiments}. The selection here is over trajectories, not over
tasks: which tasks exist was settled in Phase~2 by the feasibility check of
Sec.~\ref{sec:phase2}, and a task the teacher fails still stands, it simply
contributes no demonstration. Appendix~\ref{app:corpus} gives the per-world
composition of the resulting corpus. This is verifier-filtered,
rejection-sampled distillation \citep{star2022,rft2023}, and it inherits the
teacher's ceiling. Sec.~\ref{sec:rl} addresses that ceiling.

\section{The environment suite}
\label{sec:suite}

\subsection{Ten deep domains}

\begin{figure}[p]
  \centering
  {\captionof{table}{The ten full-domain worlds, grouped by the work they
  represent. Each is named for the workflow rather than for any product.}%
  \label{tab:suite}}
  {\small
  \begin{tabular}{@{}p{3.3cm}p{3.9cm}p{5.9cm}@{}}
  \toprule
  \textbf{Workflow category} & \textbf{Worlds} & \textbf{Depth the world must carry} \\
  \midrule
  Communication \& coordination & \echo{EchoMail}, \echo{EchoCalendar}, \echo{EchoChat} & Shared threads, schedules, participants, permissions, histories \\
  Technical creation \& operations & \echo{EchoML}, \echo{EchoForge} & Artefacts, configuration, dependencies, roles, multi-stage changes \\
  Regulated records \& transactions & \echo{EchoBank}, \echo{EchoCare} & Balances or records, authorisation, audit history, consequential writes \\
  Community, media \& travel & \echo{EchoForum}, \echo{EchoTunes}, \echo{EchoStay} & Persistent preferences, social state, search, booking, account actions \\
  \bottomrule
  \end{tabular}}

  \vspace{14pt}

  \envshot{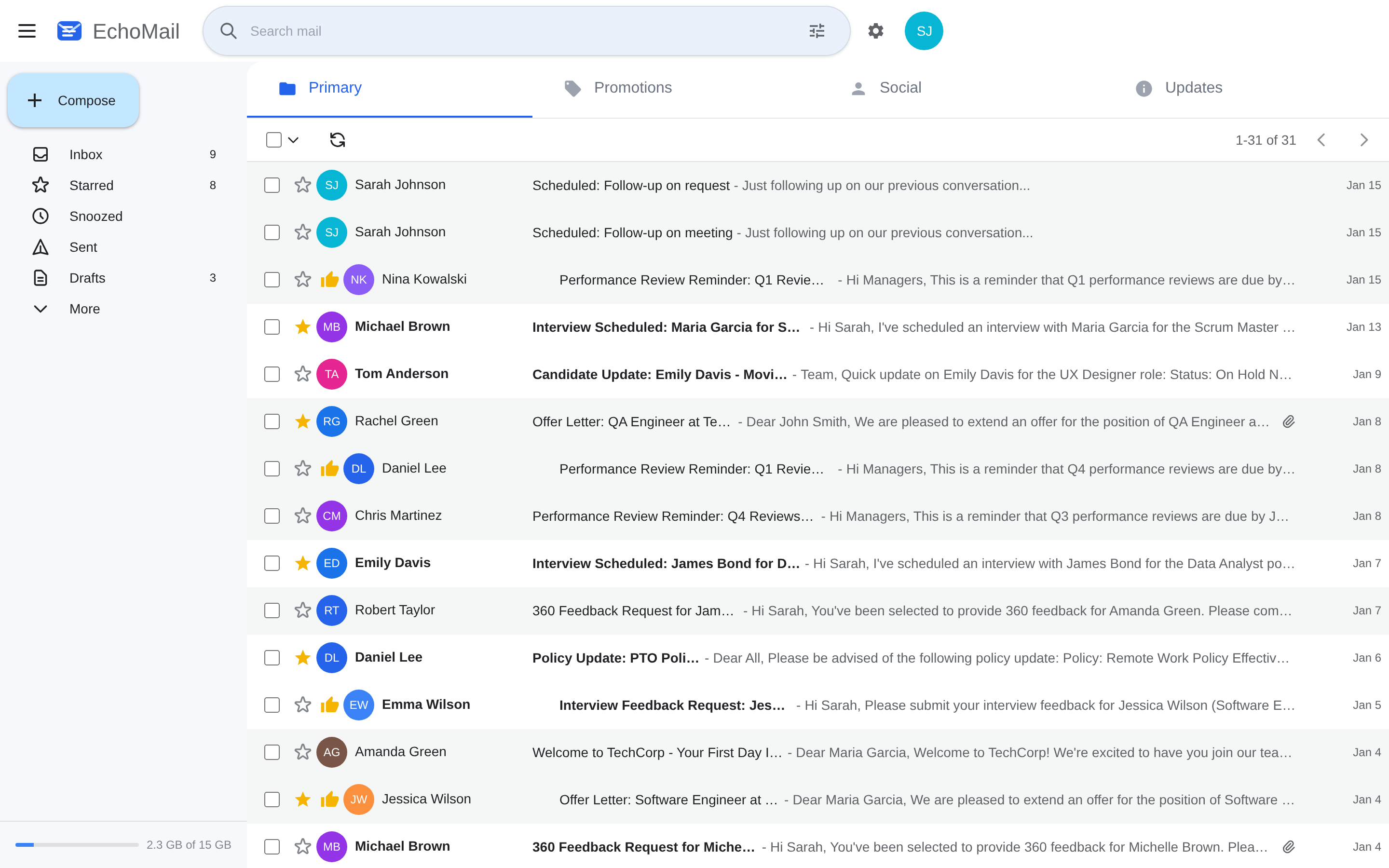}{\echo{EchoMail}} \hfill
  \envshot{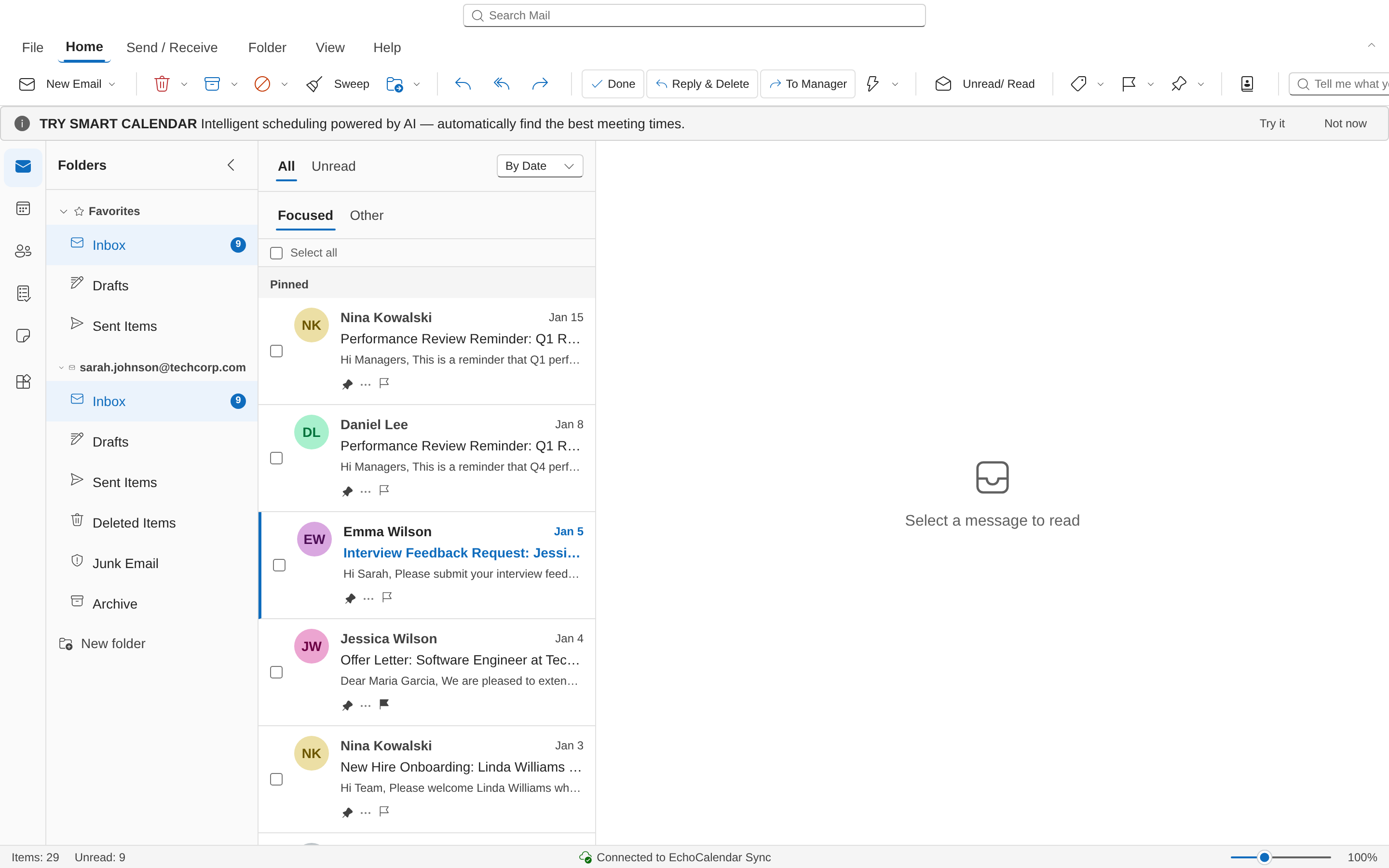}{\echo{EchoCalendar}} \hfill
  \envshot{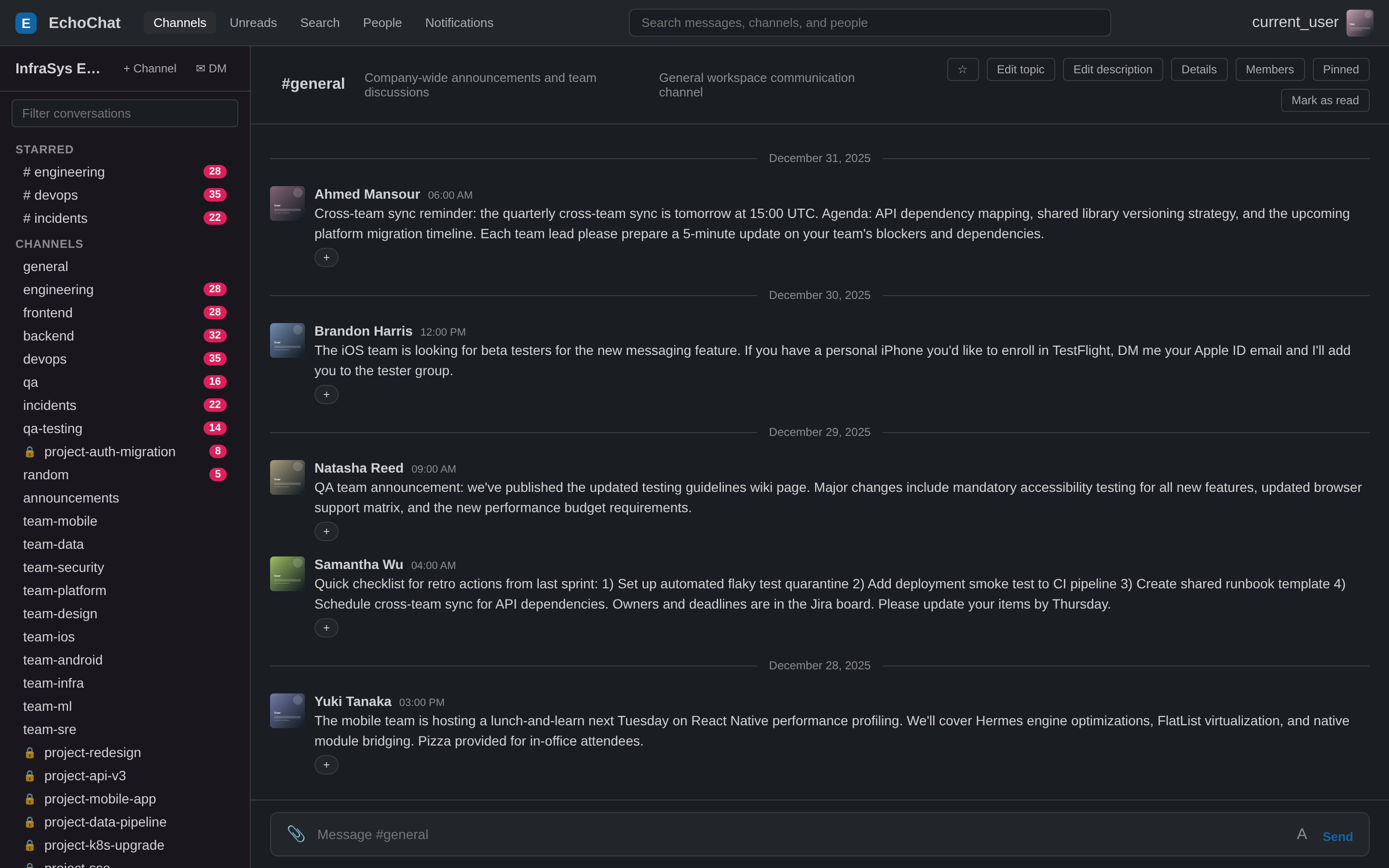}{\echo{EchoChat}}

  \vspace{4pt}
  \envshot{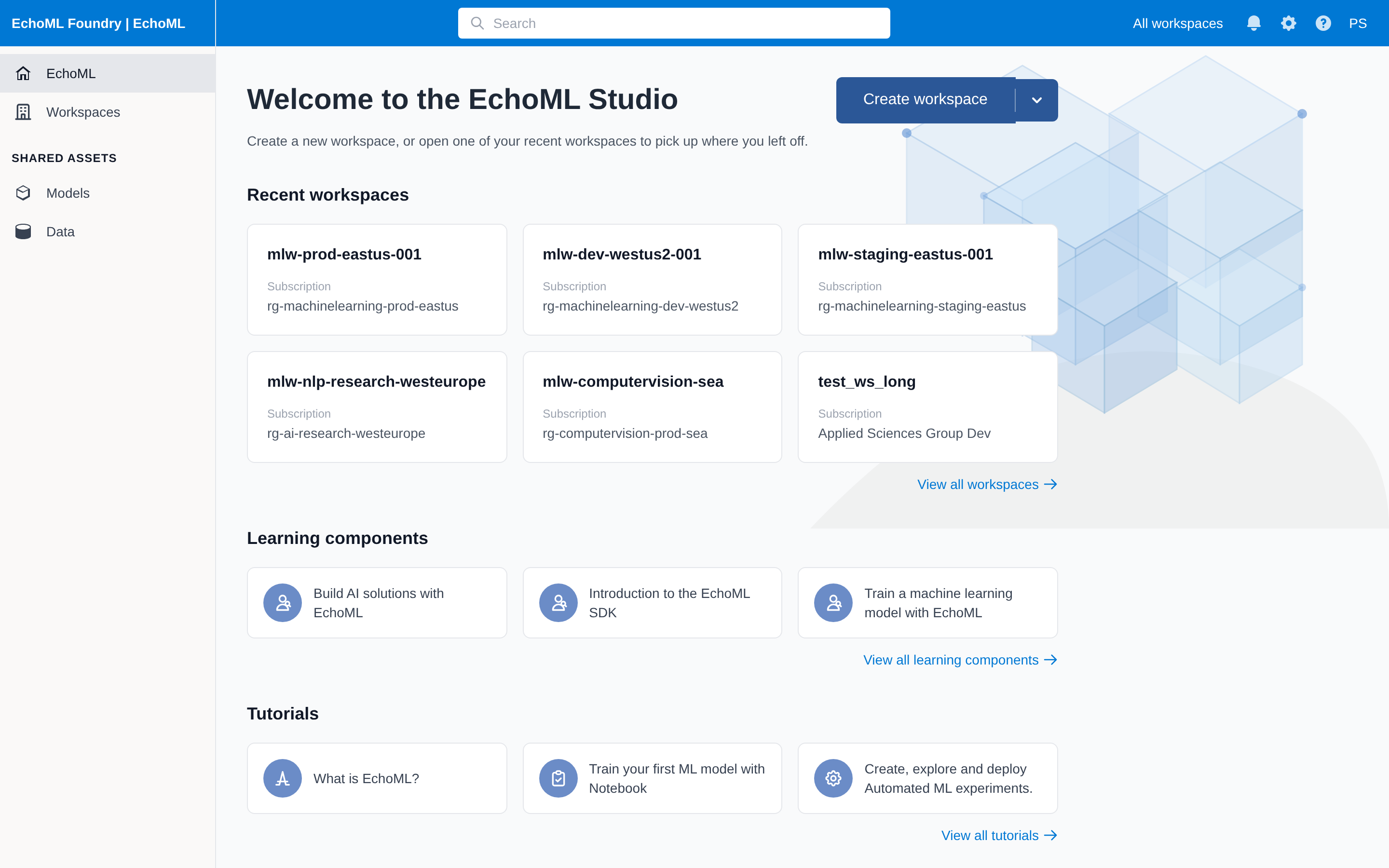}{\echo{EchoML}} \hfill
  \envshot{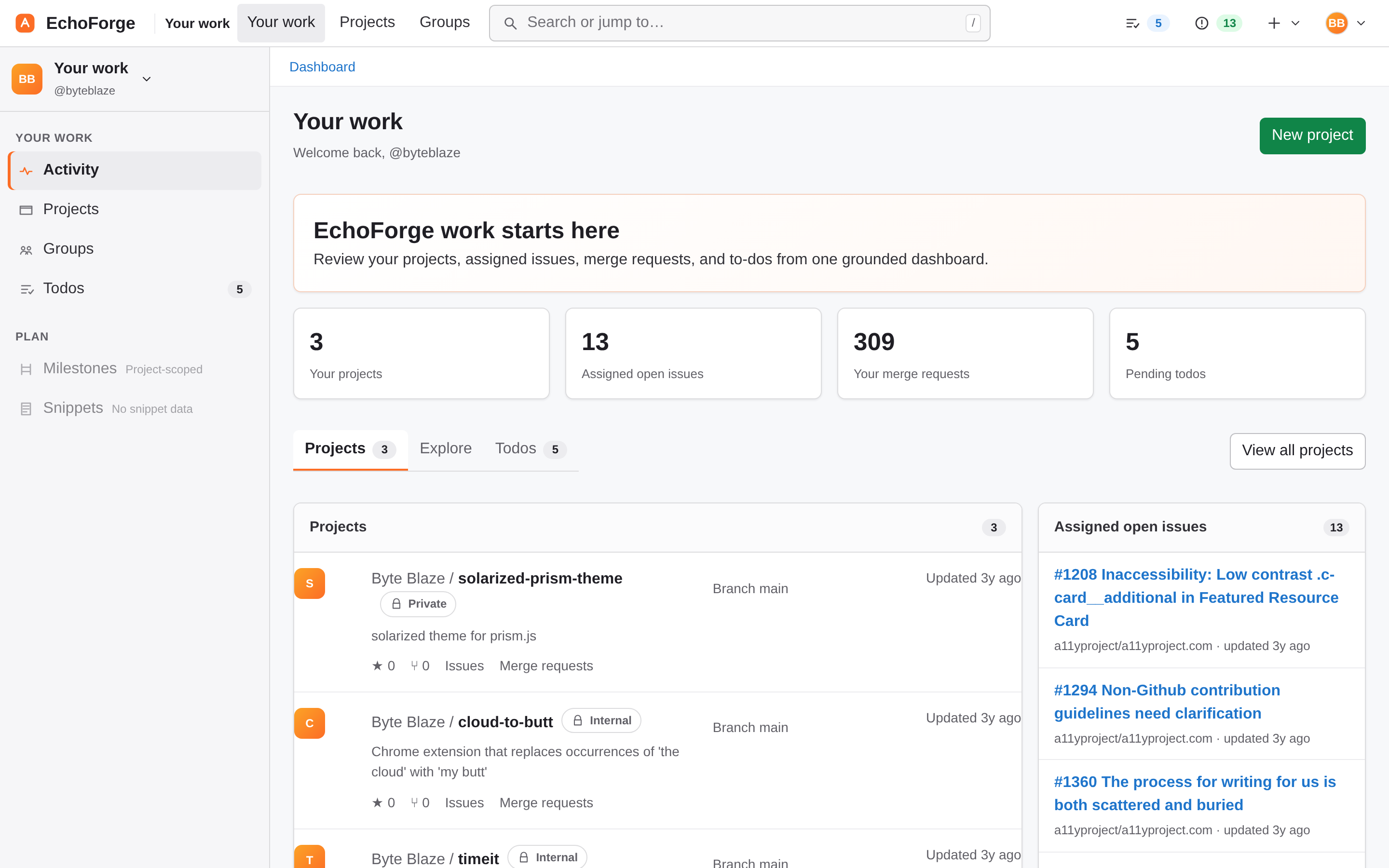}{\echo{EchoForge}} \hfill
  \envshot{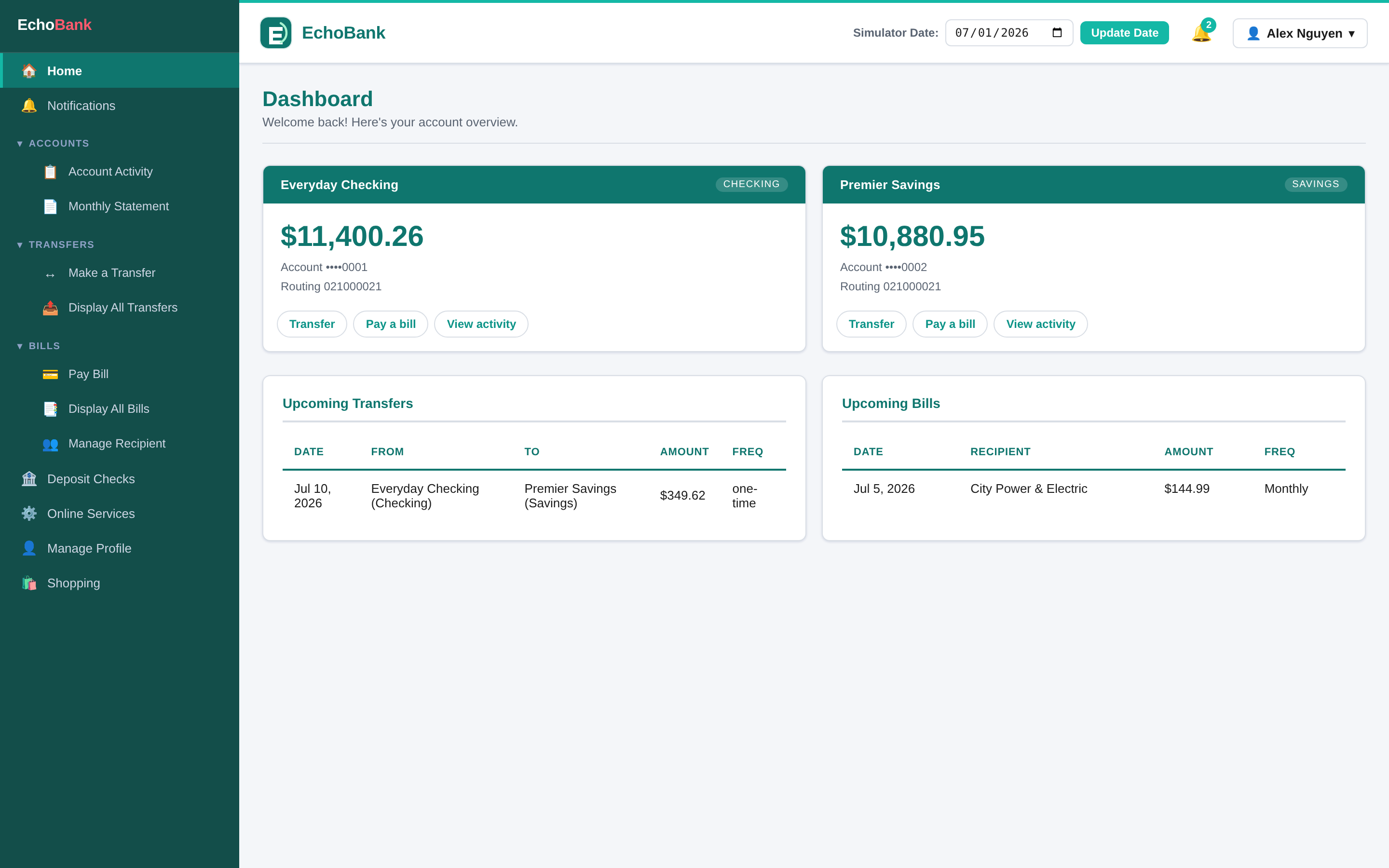}{\echo{EchoBank}}

  \vspace{4pt}
  \envshot{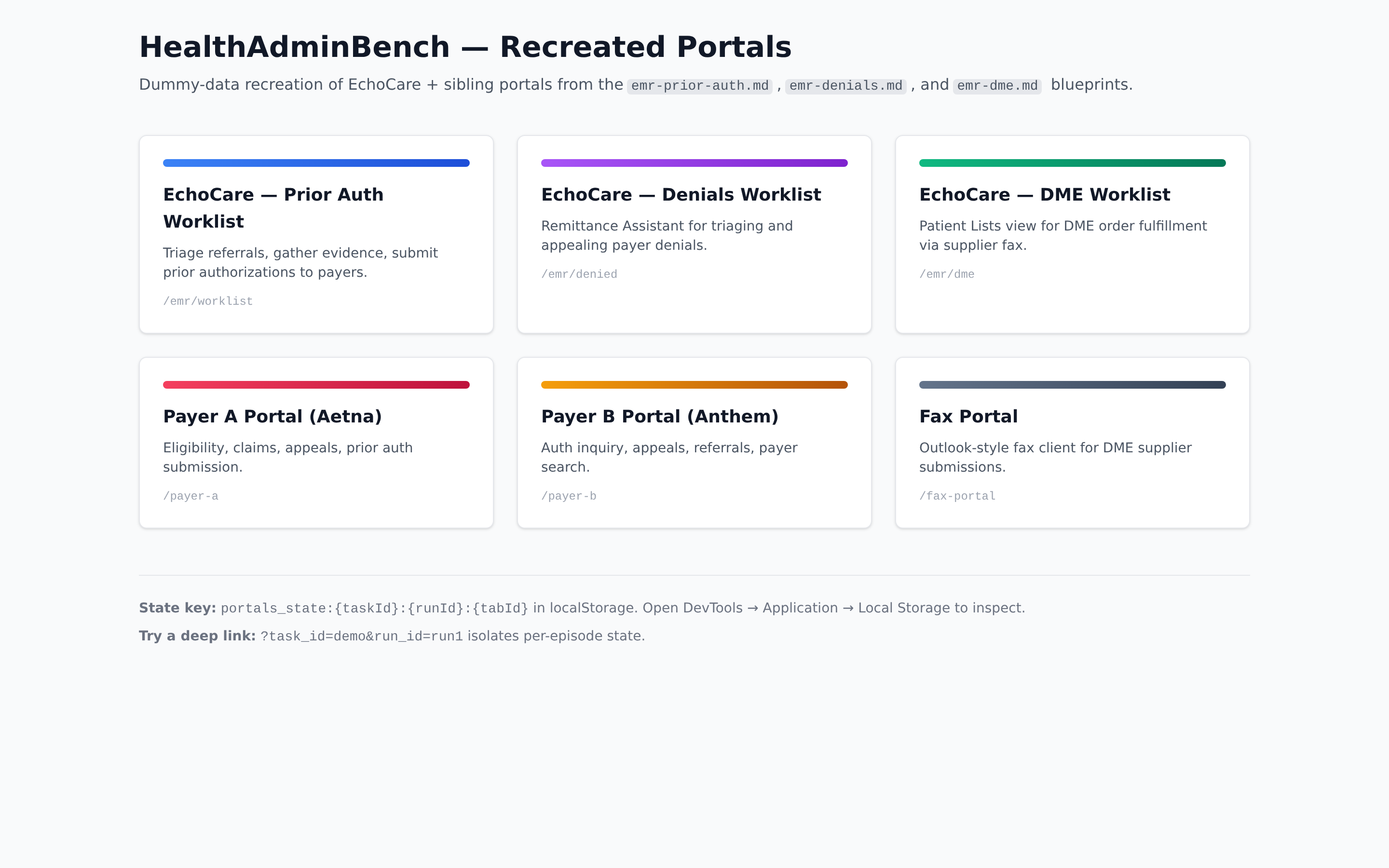}{\echo{EchoCare}} \hfill
  \envshot{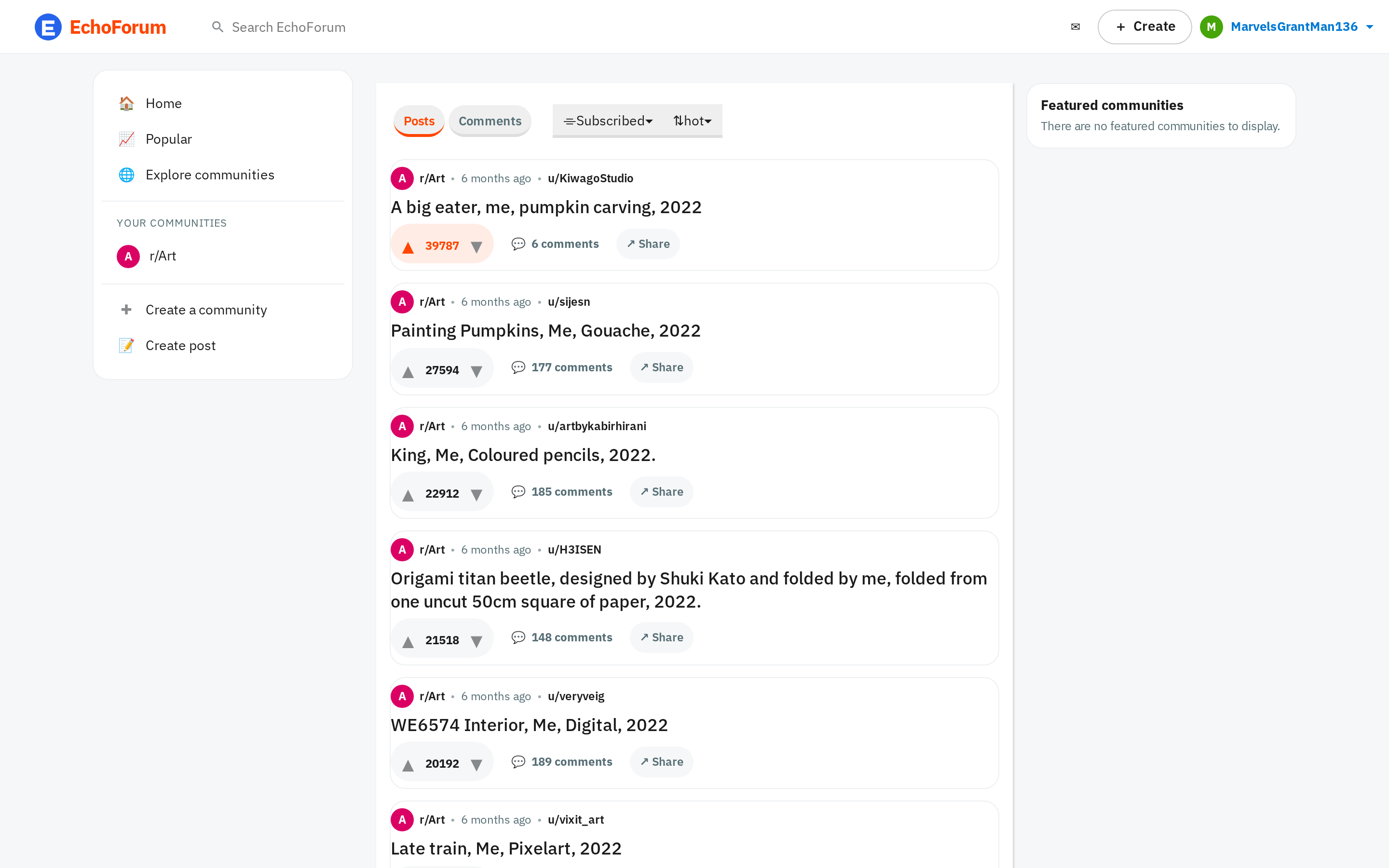}{\echo{EchoForum}} \hfill
  \envshot{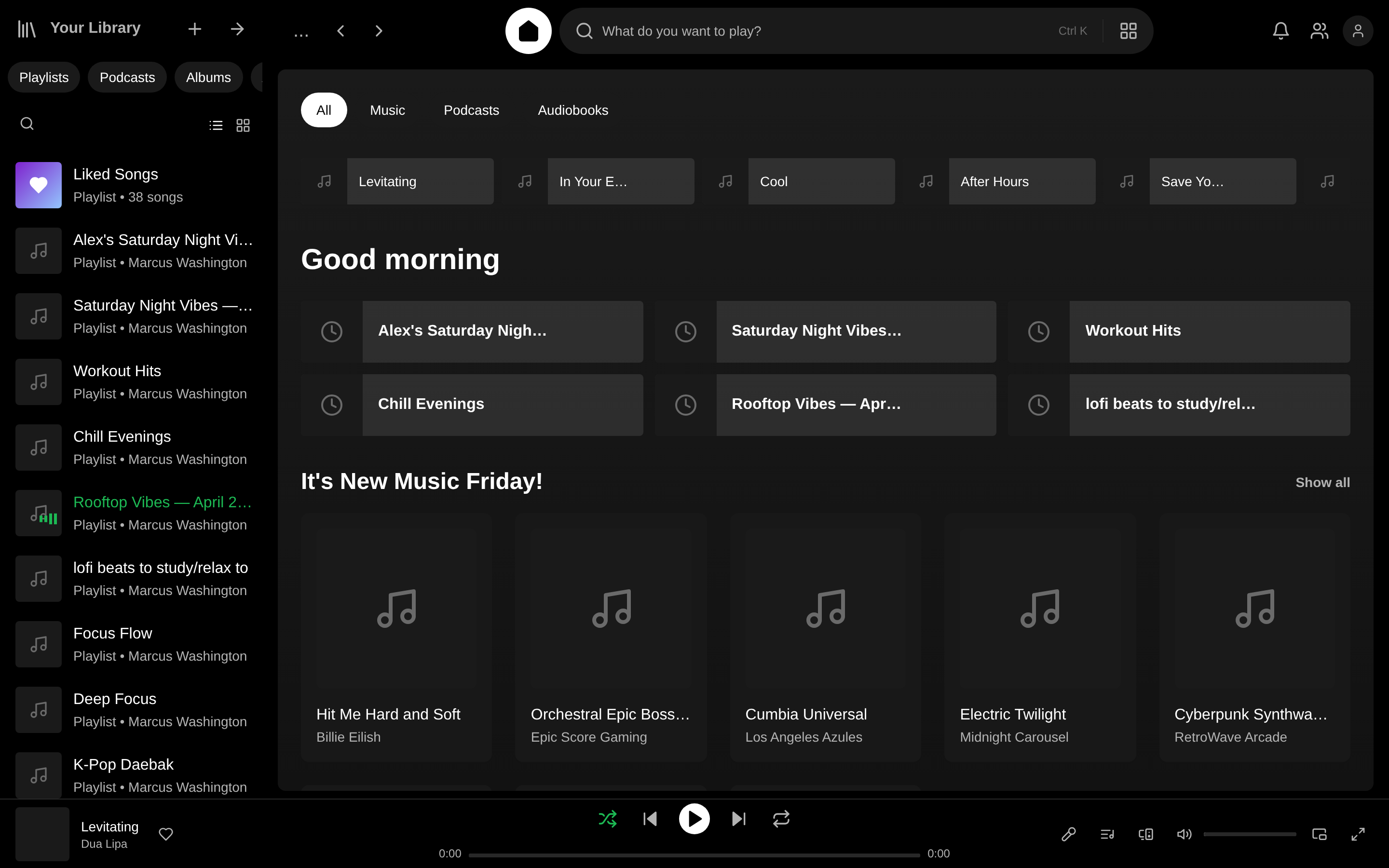}{\echo{EchoTunes}}

  \vspace{4pt}
  \envshot{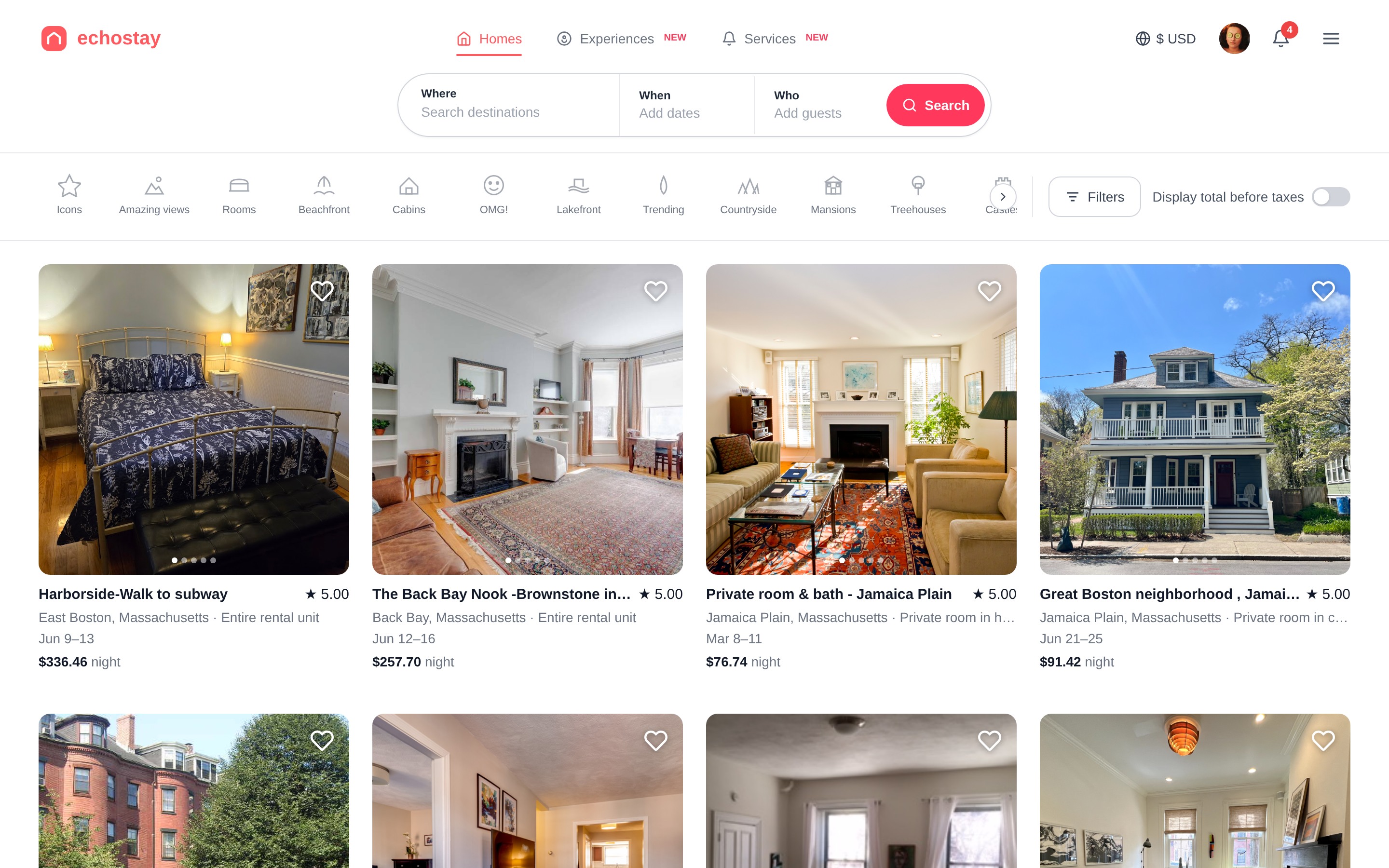}{\echo{EchoStay}} \hfill
  \envshot{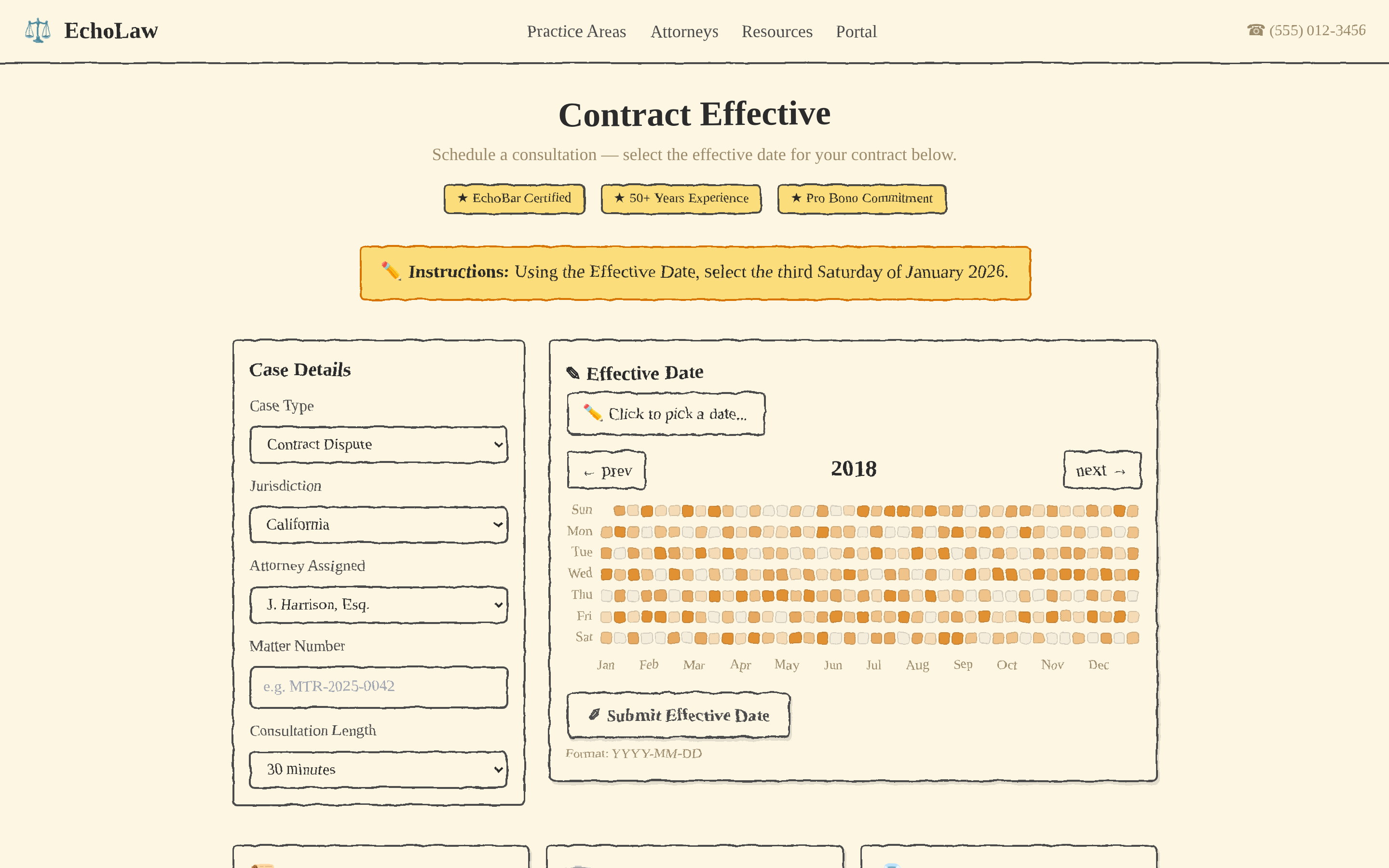}{Datepicker world, held out} \hfill
  \envshot{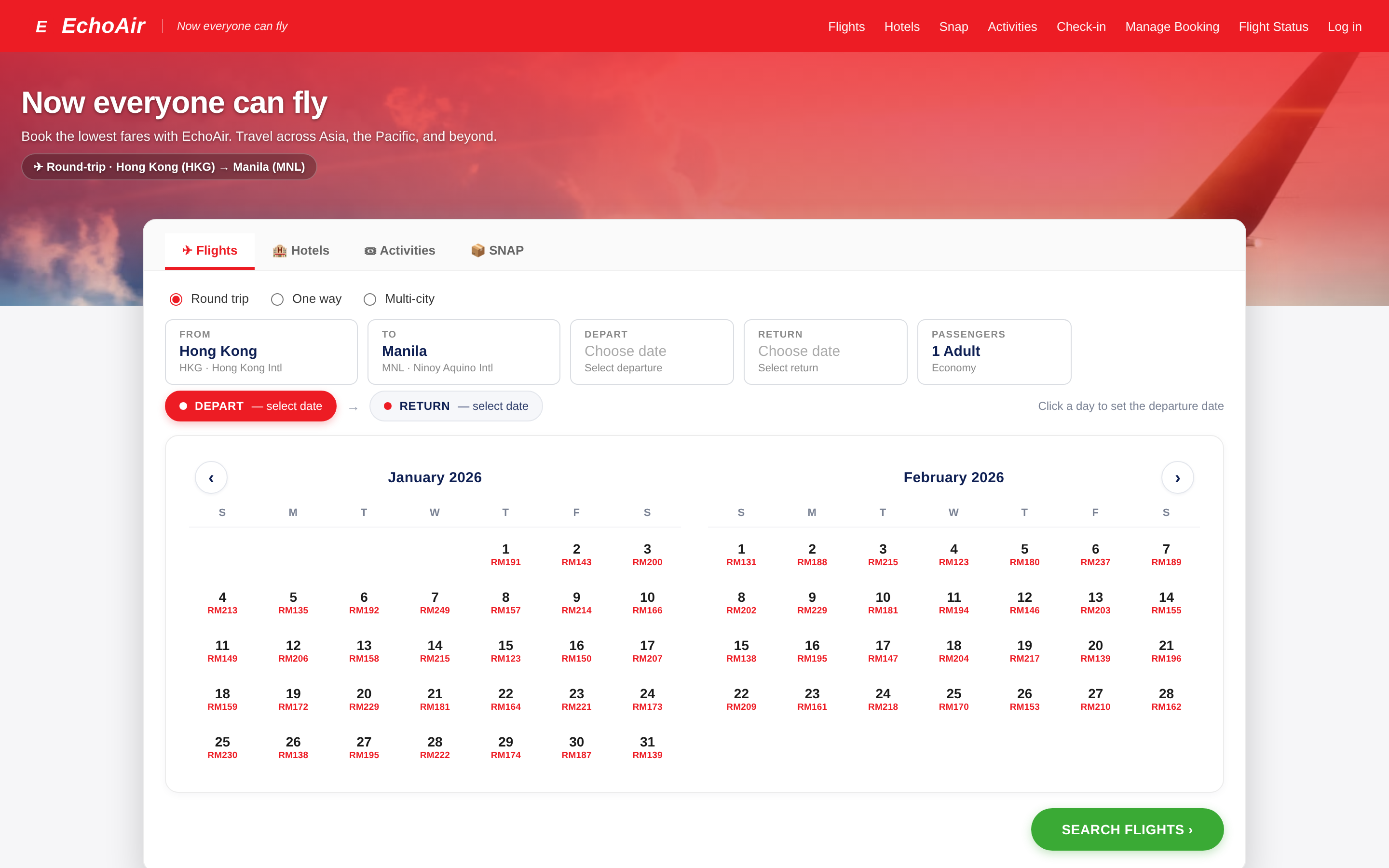}{Nested-filter world, held out}

  \caption{The suite as an agent meets it. The first ten panels are the
  full-domain worlds tabulated above, each a behaviourally faithful clone of a
  class of application rather than of any particular product. The last two are
  held-out renderings from the capability worlds of
  Sec.~\ref{sec:capability}: a calendar heatmap in a legal-services frontend and
  a compound booking-search panel in an airline frontend, neither of which
  appears in training. What a task depends on is the state behind these screens
  rather than their appearance.}
  \label{fig:suite}
\end{figure}

The domains with the most consequential work are the hardest for public
benchmarks to reach: closed, proprietary systems where the difficulty lives in
permissions, shared state and history rather than layout. What matters is not
the pixels but that an action's consequences reach across screens and users, so
that a task can run a real workflow and be graded on the state it leaves behind.
Table~\ref{tab:suite} groups the ten domains by the work they represent, and
Figure~\ref{fig:suite} shows the suite as an agent meets it.

That accumulated state is what carries consequences across screens and users. A
booking in \echo{EchoStay} moves through search, listing, availability and
payment across roughly 87 routes and 23 tables, with no confirmation screen to
short-circuit it. An \echo{EchoMail} thread carries intent from draft through
delivery, reply and label state. An \echo{EchoCare} order writes each change to
an audit trail. The tasks are expensive because of it, often five to thirty
actions deep, and finished only when the underlying state has changed.

\begin{table}[htbp]
\centering
\footnotesize
\setlength{\abovecaptionskip}{4pt}
\renewcommand{\arraystretch}{0.95}
\caption{Grounded database state for each of the ten full-domain worlds. All
counts are measured from the shipped seed database, not from a specification.}
\label{tab:suite-detail}
\begin{tabular}{@{}p{2.45cm}p{3.05cm}rrp{4.1cm}@{}}
\toprule
\textbf{World} & \textbf{Class of application} & \textbf{Tables} & \textbf{Rows} & \textbf{Principal seeded entities} \\
\midrule
\echo{EchoMail}     & Email workspace         & 13 & 5.2K & 1.1K emails, 1.1K threads, 170 labels, 374 contacts \\
\echo{EchoCalendar} & Calendar and mail       & 13 & 2.2K & 196 events, 102 folders, 374 contacts \\
\echo{EchoChat}     & Team messaging          & 15 & 34.7K & 188 channels, 1.3K workspaces, 3.2K users \\
\echo{EchoML}       & ML platform             & 11 & 3.3K & 173 models, 444 datasets, 152 endpoints, 309 compute resources \\
\echo{EchoForge}    & Developer collaboration & 15 & 705K & 175 projects, 81K issues, 134K merge requests \\
\echo{EchoBank}     & Online banking          & 10 & 1.8K & 100 accounts, 4.1K transactions, 201 transfers, 672 bills \\
\echo{EchoCare}     & Health revenue cycle    & 14 & 2.8K & 250 prior authorisations, 250 denials, 200 orders, 1.3K documents \\
\echo{EchoForum}    & Social forums           & 11 & 4.0M & 95 forums, 127K posts, 2.55M comments, 662K users \\
\echo{EchoTunes}    & Music streaming         & 42 & 21.5K & 2.5K songs, 1.1K albums, 863 artists, 544 playlists \\
\echo{EchoStay}     & Lodging marketplace     & 23 & 180K & 3.7K listings, 5.4K bookings, 5.3K reviews, 6.1K users \\
\bottomrule
\end{tabular}
\end{table}

\subsection{Two capability worlds}
\label{sec:capability}

Not every weakness is a missing domain. Some are a single control the agent
cannot reliably operate. Consider an agent that searches, filters and opens the
right listing, then stalls at the date picker, unable to turn ``the second week
of March'' into the right clicks on an unfamiliar calendar. Building another
booking site would not fix that. The skill is learned only when the control
itself appears in enough forms, and date pickers and nested filter-and-search
controls are rendered a hundred different ways on the live web, which is
variability a single deep application cannot supply.

\begin{figure}[tbp]
  \centering
  \includegraphics[width=0.82\linewidth]{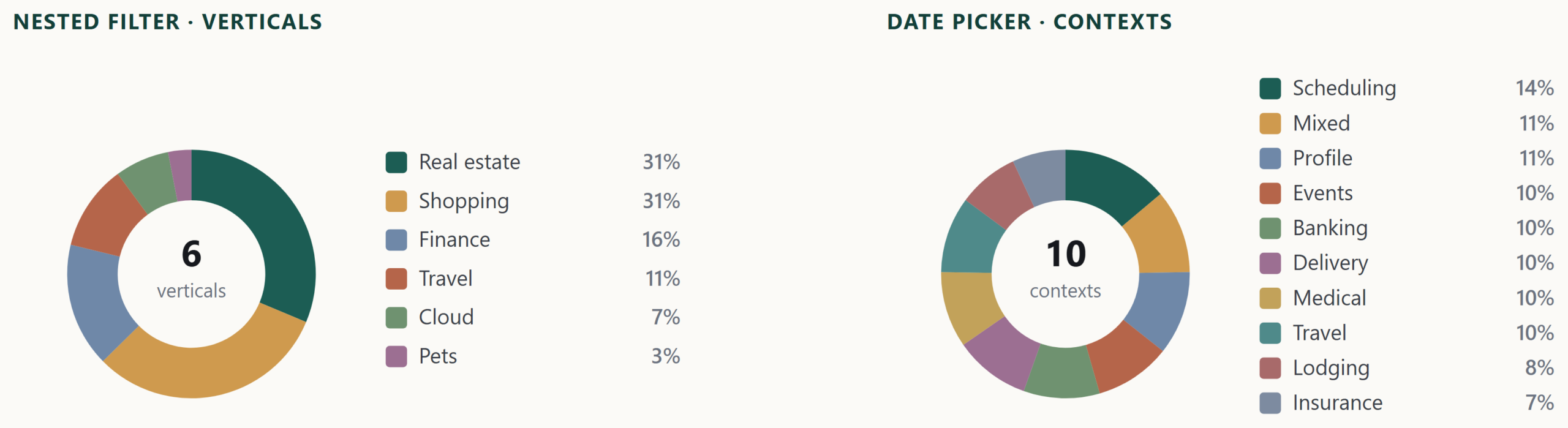}
  \caption{Thematic spread of the two capability worlds. The same control is
  re-themed so that the skill cannot be learned as a single layout: nested
  filters over six verticals, date pickers over ten contexts. Percentages are
  shares of generated frontends within each world.}
  \label{fig:coverage}
\end{figure}

We therefore isolate the control and widen the interaction, mass-producing it
across layouts, states and constraints, then generating grounded tasks over
each. The datepicker world renders one date control as six core widgets across
ten contexts, 100 frontends in six visual styles, and holds out ten unseen
widgets over 36 further scenarios, 80 frontends in nine styles, from calendar
heatmaps to scroll wheels and fiscal-quarter pickers. Its hardest tasks turn
transcription into reasoning, resolving ``the last Thursday of January 2026'' or
``10 business days after a start date'' to one exact, widget-reachable date. The
nested-filter world varies twenty widget families over 200 frontends in fifteen
styles, and holds out nine compound panel families over 100 frontends in nine
styles, grading every submission by whether the filtered results actually meet
the requested conditions, judged by the application's own logic rather than by
appearance. Neither world lets a single theme dominate, since a control learned
only against one vertical is a layout rather than a skill;
Figure~\ref{fig:coverage} gives the thematic spread of both, and
Appendix~\ref{app:widgets} lists every family with the interaction it demands.

\subsection{Twelve training worlds, fourteen evaluation splits}

To avoid ambiguity in the tables that follow: training uses \textbf{twelve
worlds}, the ten full domains plus the two capability worlds, while evaluation
reports \textbf{fourteen splits}, the ten domains plus an in-distribution and a
held-out split for each capability world. All averages in this paper are
unweighted means over the fourteen splits.

\newlength{\origtextfloatsep}\setlength{\origtextfloatsep}{\textfloatsep}
\newlength{\origintextsep}\setlength{\origintextsep}{\intextsep}
\setlength{\textfloatsep}{9pt plus 2pt minus 2pt}
\setlength{\intextsep}{9pt plus 2pt minus 2pt}

\section{Supervised experiments}
\label{sec:experiments}

\subsection{Setup}
\label{sec:exp-setup}

Three models run through this section: \textbf{Base} ($\pibase$), Qwen3.5-9B
\citep{qwen35} given only enough synthetic trajectories to align it to the
browser action space; \textbf{Ours} ($\pisft$), that same 9-billion-parameter
network trained on the full corpus of Sec.~\ref{sec:sft}, 21{,}009 verified
trajectories over the twelve worlds (Appendix~\ref{app:corpus}); and
\textbf{GPT-5.4}, the far larger frontier model that generated it.

Policies are served with vLLM \citep{vllm2023}. Every rollout is capped at 100
actions, and a task that reaches the cap without emitting \texttt{terminate} is
recorded as over-budget. Tasks in the synthetic worlds are graded by the
database-grounded verifiers of Sec.~\ref{sec:grading}, while WebVoyager and
Online-Mind2Web use their standard published judges. Live-web runs are reported
both through Browserbase \citep{browserbase}, which strips the datacentre
bot-blocks and rate limits that depress every agent's score, and without it.

\subsection{Deep worlds transfer; shallow worlds set the model back}
\label{sec:exp-depth}

A shallow world is the cheap option: it stands up fast and looks convincing, but
it rehearses only isolated, correct-looking clicks, so a model trained on it
picks up habits the easy world never punished, over-stepping, looping and
repeating dead actions. A deep world costs more, but its trajectories carry the
dependent structure that transfers.

To isolate this we take two live WebVoyager domains, Allrecipes and Hugging
Face, and compare three checkpoints: $\pibase$ and two ablation models,
$\pishallow$ and $\pideep$, trained on shallow-world and deep-world trajectories
built for those two domains alone, so neither is $\pisft$. The shallow
world poses short, self-contained tasks; the deep world poses tasks that run
across dependent steps, where an early action changes the state, options and
verification available later. Both see the same two domains and corpora
of comparable size, though not matched trajectory for trajectory. Evaluation
uses public WebVoyager tasks for these domains, run on the live sites outside
any training world.

\begin{figure}[htbp]
  \centering
  \includegraphics[width=0.62\linewidth]{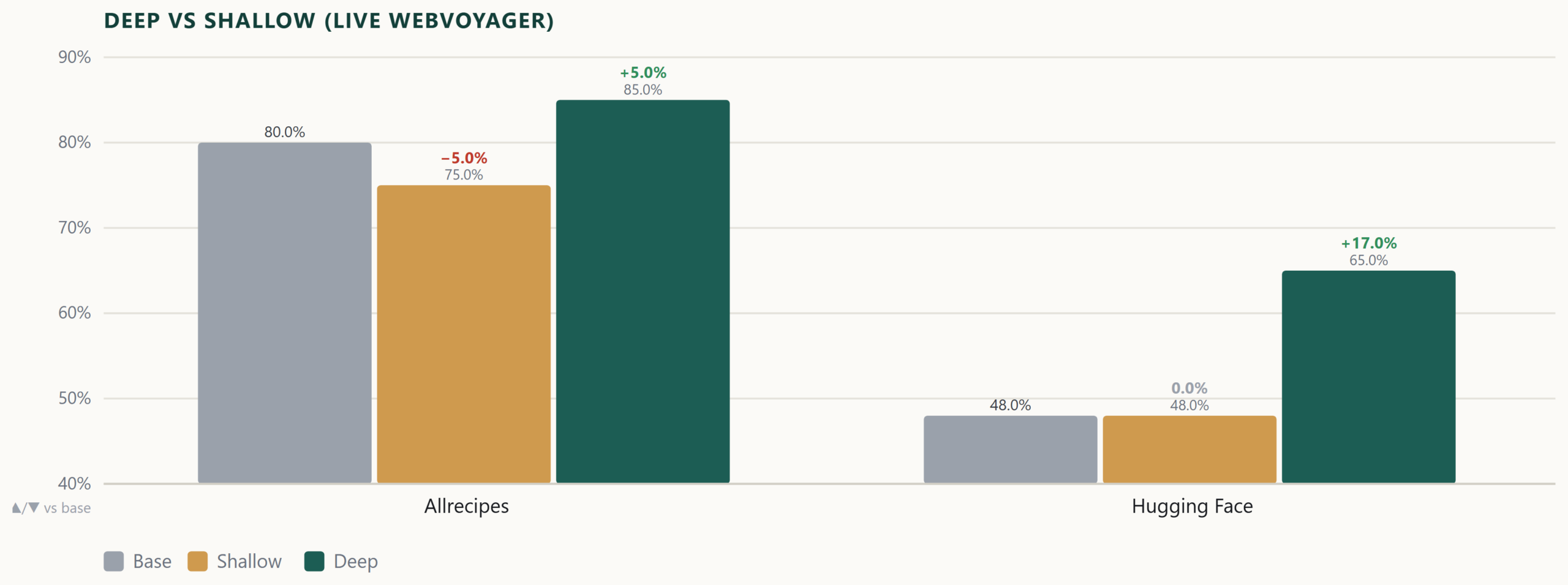}
  \caption{Deep versus shallow worlds on two live WebVoyager domains, with the
  same domain coverage and differing task depth. Deep lifts both; shallow
  drops below base on Allrecipes and stalls on Hugging Face.}
  \label{fig:depth}
\end{figure}

On Allrecipes $\pishallow$ falls below base, $80.0 \to 75.0$; on
Hugging Face it stays flat at $48.0$. Only $\pideep$ improves both,
lifting Allrecipes to $85.0$ and the harder Hugging Face split to $65.0$
(Figure~\ref{fig:depth}). Since both runs cover the same domains at comparable
scale, depth is the systematic difference between them: what
separates improvement from regression is not how much the model saw but whether
what it saw preserved the structure of the work. The mechanism shows in the
trajectories: $\pideep$ loops less, and of the 37 Hugging Face tasks those
that exhaust their step budget fall from 15 to 9. With two domains and a few
dozen tasks each, we read this as evidence that a shallow world can be worse
than no training at all, not as an estimate of how much depth is worth.

\clearpage
\subsection{The full scorecard: base, our model and a frontier reference}
\label{sec:exp-scorecard}

This is the main supervised result: what the twelve worlds, through the
21{,}009-trajectory corpus of Appendix~\ref{app:corpus}, do to a single 9B
model. Three policies run the same fourteen splits under the same harness, the
base model $\pibase$, our supervised model $\pisft$ trained on that corpus, and
GPT-5.4 as frontier reference and as the teacher that produced it. Every entry
is task success under the
database-grounded verifiers of Sec.~\ref{sec:grading}, not a judgement about
appearance. Table~\ref{tab:scorecard} lists the splits;
Figure~\ref{fig:scorecard} plots the same numbers as distance still to travel.
Across all fourteen splits $\pisft$ nearly doubles base, $36.5 \to 67.1$, and on
the six splits where base was weakest it climbs three- to nine-fold, from single
or low double digits into the forties through sixties. That leaves a 9B model
within fourteen points of GPT-5.4 on the average, matching or beating it on
\echo{EchoMail}, \echo{EchoBank} and both nested-filter splits, on training data
that is deep, targeted and checkable rather than on scale.

\begin{table}[!ht]
\centering
\footnotesize
\setlength{\abovecaptionskip}{4pt}
\renewcommand{\arraystretch}{0.9}
\caption{Task success (\%) over all fourteen evaluation splits, ordered by
$\pisft$. Bold marks the splits where the 9B model matches or beats GPT-5.4.}
\label{tab:scorecard}
\begin{tabular}{@{}lrrr@{}}
\toprule
\textbf{Split} & \textbf{Base} $\pibase$ & \textbf{Ours} $\pisft$ & \textbf{GPT-5.4} \\
\midrule
\echo{EchoBank}                & 76.6 & \textbf{94.6} & 92.8 \\
Datepicker (in-distribution)   & 60.0 & 87.2 & 89.8 \\
Nested filter (in-distribution)& 76.0 & \textbf{87.0} & 87.0 \\
Nested filter (held out)       & 62.8 & \textbf{84.1} & 82.8 \\
\echo{EchoMail}                & 64.2 & \textbf{81.1} & 74.5 \\
\echo{EchoTunes}               & 35.9 & 70.6 & 81.7 \\
\echo{EchoCalendar}            & 13.6 & 66.9 & 78.1 \\
\echo{EchoForum}               & 20.4 & 62.0 & 88.7 \\
\echo{EchoCare}                & 15.2 & 61.6 & 78.4 \\
\echo{EchoML}                  &  6.5 & 56.0 & 80.6 \\
\echo{EchoForge}               & 16.8 & 52.5 & 74.3 \\
Datepicker (held out)          & 34.0 & 52.0 & 94.7 \\
\echo{EchoChat}                &  8.6 & 46.7 & 76.2 \\
\echo{EchoStay}                & 20.5 & 37.6 & 50.4 \\
\midrule
\textbf{Average}               & \textbf{36.5} & \textbf{67.1} & \textbf{80.7} \\
\bottomrule
\end{tabular}
\end{table}

\begin{figure}[!ht]
  \centering
  \setlength{\abovecaptionskip}{4pt}
  \includegraphics[width=\linewidth]{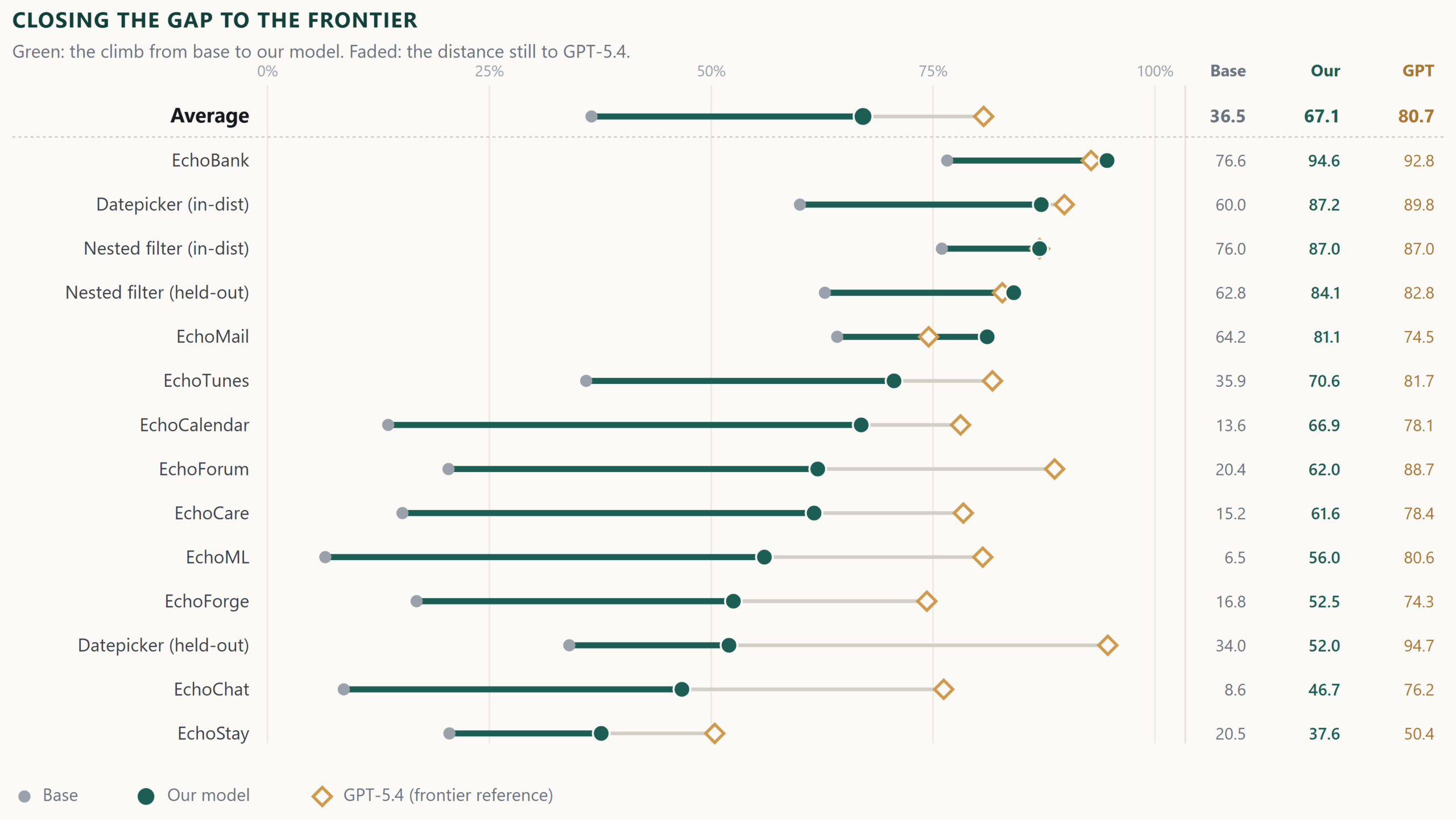}
  \caption{Closing the gap to the frontier, per evaluation split. The solid bar
  is the climb from $\pibase$ to $\pisft$; the faded remainder is the distance
  still to GPT-5.4.}
  \label{fig:scorecard}
\end{figure}

\clearpage
\setlength{\textfloatsep}{\origtextfloatsep}
\setlength{\intextsep}{\origintextsep}
\subsection{Precision about one skill}
\label{sec:exp-capability}

The datepicker and nested-filter worlds drill the controls our evaluations
flagged, and the two skills reinforce each other rather than competing. The four
checkpoints below are capability-only ablations, trained on those two worlds
rather than on the full corpus, so none of them is $\pisft$ and their scores are
not the ones Table~\ref{tab:scorecard} reports.

\begin{figure}[!ht]
  \centering
  \includegraphics[width=\linewidth]{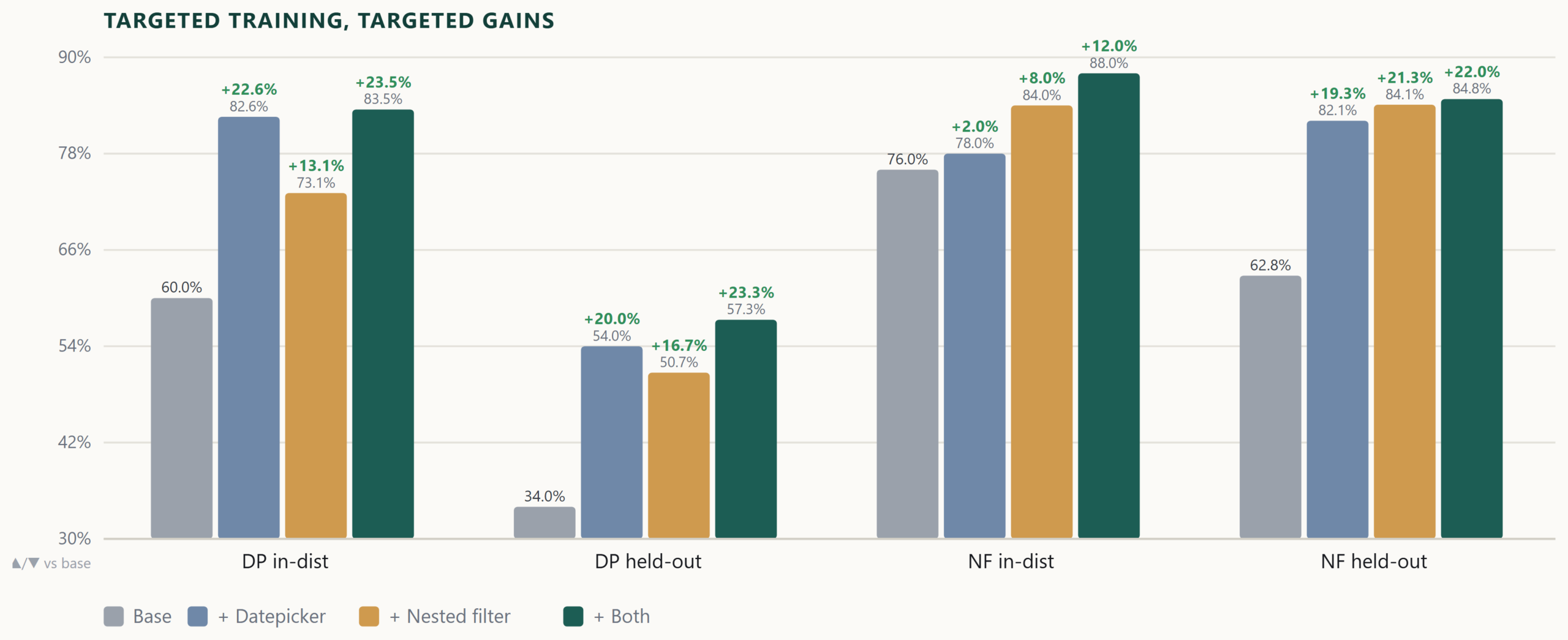}
  \caption{Targeted training, targeted gains. Training either skill lifts both
  controls, including held-out widgets and compositions neither was trained on,
  and training both is the best all-rounder on every split.}
  \label{fig:capability}
\end{figure}

Two findings matter, both visible in Table~\ref{tab:capability} and summarised
in Figure~\ref{fig:capability}. First, gains hold on forms never trained on,
with held-out
date pickers rising $34.0 \to 57.3$ and held-out filter panels $62.8 \to 84.8$,
which indicates these models learned a rule rather than a layout. Second, the
skills transfer across each other rather than cannibalising: training either
alone still lifts the other (datepicker training lifts held-out filters to
$82.1$, filter training lifts held-out date pickers to $50.7$), and training
both is the best all-rounder on every split. Against GPT-5.4 as a frontier
reference, the $+$Both model edges ahead on nested filters and closes most of
the datepicker in-distribution gap, trailing clearly only on held-out date
pickers, where GPT-5.4 is exceptionally strong ($94.7$). The effect also reaches
the open web: $+$Both lifts Online-Mind2Web from $29.5$ to
$34.3$ on sites it never saw.

\begin{table}[!ht]
\centering
\small
\caption{Capability ablation: four models over four splits (task success, \%).
Best per row in bold. ``Held out'' widget families never appear in training.}
\label{tab:capability}
\begin{tabular}{@{}lrrrr@{}}
\toprule
\textbf{Split} & \textbf{Base} & \textbf{$+$Datepicker} & \textbf{$+$Nested filter} & \textbf{$+$Both} \\
\midrule
Datepicker, in-distribution  & 60.0 & 82.6 & 73.1 & \textbf{83.5} \\
Datepicker, held out         & 34.0 & 54.0 & 50.7 & \textbf{57.3} \\
Nested filter, in-distribution & 76.0 & 78.0 & 84.0 & \textbf{88.0} \\
Nested filter, held out      & 62.8 & 82.1 & 84.1 & \textbf{84.8} \\
\midrule
Mean & 58.2 & 74.2 & 73.0 & \textbf{78.4} \\
\bottomrule
\end{tabular}
\end{table}

\FloatBarrier
\subsection{Transfer to the live web}
\label{sec:exp-liveweb}

To test whether the skill survives the open web, we evaluate $\pisft$,
unchanged, on WebVoyager and Online-Mind2Web, benchmarks it never trained on.
These barely overlap with what we built, since both are dominated by open,
public sites and read-mostly browsing while our worlds train login-gated,
write-heavy workflows. We therefore expect direction rather than a large jump.
Table~\ref{tab:liveweb} reports both browser conditions.

\begin{table}[htbp]
\centering
\small
\caption{Live-web transfer (task success, \%). Neither benchmark contributes any
training data. Hosted rows run through a browser service that removes datacentre
bot-blocks; datacentre rows do not.}
\label{tab:liveweb}
\begin{tabular}{@{}llrr@{}}
\toprule
\textbf{Benchmark} & \textbf{Browser} & \textbf{Base} $\pibase$ & \textbf{Ours} $\pisft$ \\
\midrule
WebVoyager        & hosted     & 66.5 & 71.5 \\
WebVoyager        & datacentre & 50.9 & 55.6 \\
Online-Mind2Web   & hosted     & 40.5 & 43.4 \\
Online-Mind2Web   & datacentre & 29.5 & 37.2 \\
\bottomrule
\end{tabular}
\end{table}

With no live-web data in the mix, this is transfer rather than memorisation. The
modest size of the gain reflects coverage rather than a ceiling, and aiming at a
live domain grows it. \echo{EchoForge}, our code-hosting world, is the same kind
of application as GitHub, one of the live sites WebVoyager tests, and adding it
to an earlier, smaller training mix lifted that mix's live GitHub score from
$58.5$ to $63.4$, with its overall live scores rising too (WebVoyager
$50.9 \to 52.9$, Online-Mind2Web $29.5 \to 31.1$); those figures are that
ablation's, not $\pisft$'s. The average reflects that most of what we built sits
in domains these benchmarks never touch.

\subsection{What scaling buys, and what it does not}
\label{sec:exp-scaling}

\begin{figure}[htbp]
  \centering
  \includegraphics[width=\linewidth]{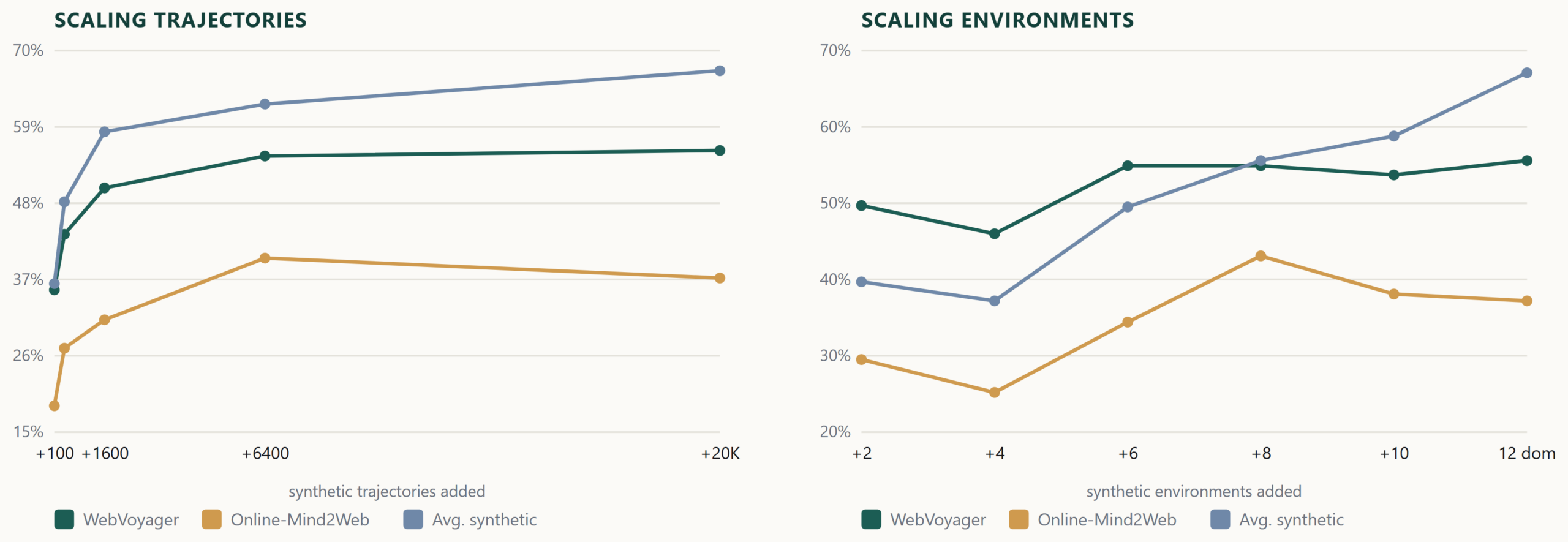}
  \caption{Two scaling axes, scored without a hosted browser. Left: more
  trajectories over a fixed set of worlds, subsampled from the full corpus at
  its own per-world proportions, with the
  horizontal axis spaced by actual trajectory count. The synthetic average keeps
  rising, but live-web transfer saturates. Right: more environments, where
  breadth keeps both the synthetic average and WebVoyager climbing.}
  \label{fig:scaling}
\end{figure}

We scaled two axes separately: more trajectories through a fixed set of
environments, subsampled from the 21{,}009-trajectory corpus of
Appendix~\ref{app:corpus} at its own per-world proportions, and more distinct
environments. They behave differently, as Figure~\ref{fig:scaling} shows.

More trajectories on the same worlds keep lifting the in-domain average, though
the gains shrink, while transfer to the live web flattens outright: from 6\,400
to 20\,000 trajectories, WebVoyager holds steady ($54.8 \to 55.6$) and
Online-Mind2Web slips ($40.1 \to 37.2$). Since every point holds the per-world
mixture fixed, this is not an artefact of composition. Each environment holds
only so
much transferable skill, and once a model has drawn it out, more rollouts polish
what it already does.

Scaling environments produces the opposite result: the average keeps climbing as
breadth grows, and WebVoyager reaches its best only with the full set. For
generalisation the lever is diversity rather than volume. Scale by itself is not
the lever on either axis, since a large trajectory budget spent on shallow
worlds, or graded against the wrong answer, moves the synthetic number and goes
nowhere on the live web.

\subsection{The model is not the only thing that learns}
\label{sec:exp-coevo}

\echo{EchoStay} made the co-evolution loop visible. Its failures traced to the
world rather than the agent: a guest-count control silently broke booking tasks,
so a correct booking could never register. Fixing it raised the share of those
bookings that could be completed at all from $48\%$ to $78\%$, recovering 15 of
the 24 that had been blocked. The loop finds different faults elsewhere.
\echo{EchoForum} needed frontend fixes and a page-load speed-up, which took one
failing set of 37 tasks from 0 solved to 36. \echo{EchoChat}'s verifier had
drifted out of sync with the data, and realigning it lifted the share of
gradable tasks from $34\%$ to $99\%$. \echo{EchoCare} needed one state-wiring
fix, and \echo{EchoForge} had the backend logic but no user-interface control to
reach it.

\begin{figure}[htbp]
  \centering
  \includegraphics[width=0.75\linewidth]{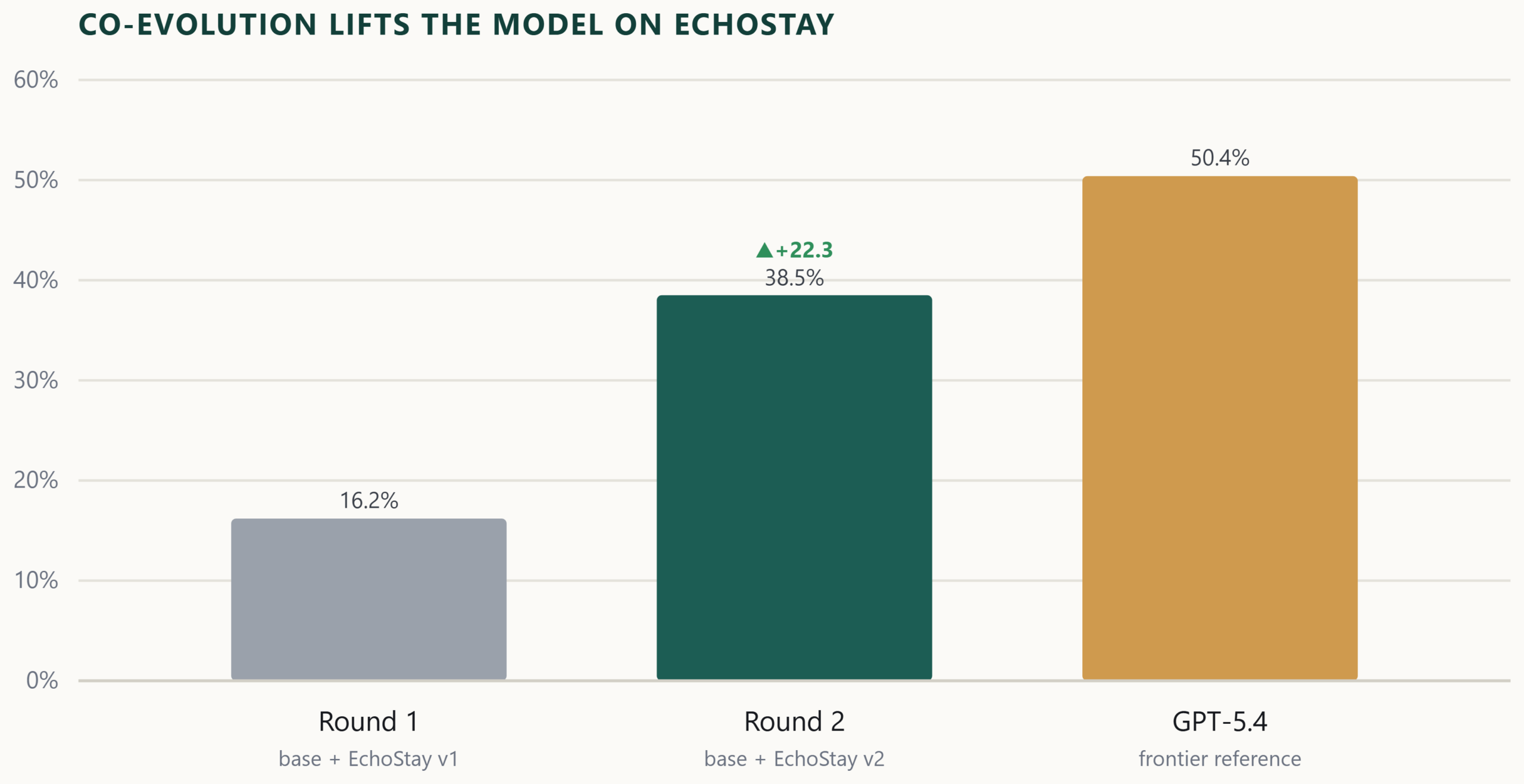}
  \caption{Co-evolution lifts the model on \echo{EchoStay}. As the world went
  from v1 to v2, the model trained on its corpus more than doubled, from
  $16.2\%$ to $38.5\%$. This is a separate measurement from the world's own
  solve rate.}
  \label{fig:coevo}
\end{figure}

As the world sharpens, the model climbs with it (Figure~\ref{fig:coevo}).
Re-running the loop on
\echo{EchoStay} across two rounds, the model trained on its corpus more than
doubles, from $16.2\%$ to $38.5\%$, two-thirds of the distance to GPT-5.4's
$50.4\%$.

The environment, the corpus and the verifier move together here, and that is
what co-evolution is rather than a confound we could have controlled away.
Which tasks exist is a function of the environment. A task is grounded on live
state and exported only if the feasibility check of Sec.~\ref{sec:phase2} finds
it reachable through the running interface, so a booking flow that cannot set a
guest count cannot yield a task that asks for one. Repairing that control does
not merely make existing tasks easier: it brings tasks into existence that could
not previously be posed, and the references those tasks are graded against are
minted from the repaired database. The two rounds are therefore each measured in
their own world, against the tasks that world can pose. An environment-only or
verifier-only ablation is not well defined under this design, and what the
comparison reports is a round of the loop as a whole rather than the effect of
any one repair.

\paragraph{The boundary the live web erases.} Inside a controlled environment
the distinction between repairing the world and teaching the model is easy to
hold, because both are inspectable. The live web erases it. There is no world
to repair mid-task, so when an action lands on nothing, correctness rests
entirely on the agent noticing and choosing differently. That is the last thing
a world has to teach, and it is where imitation is weakest.

\section{Reinforcement learning on the worlds}
\label{sec:rl}

\subsection{Imitation carries a ceiling}

Every result so far comes from imitation: the 9B model copies the trajectories
GPT-5.4 got right. Filtering on the verifier is what makes that corpus clean,
and cleanliness is the one thing it cannot teach. A trajectory that only ever
goes right contains no instance of a wrong turn being noticed and undone, and
none of an agent judging that it is finished. Sec.~\ref{sec:exp-coevo}
identified those as the residue that survives once the world itself is correct.

Keeping the failed trajectories would not close the gap either, because the
failures would still be the teacher's. A frontier model goes wrong in ways a 9B
policy mostly does not, and the 9B policy goes wrong in ways the teacher rarely
does: misreading a control that is plainly visible, repeating an action on an
element that never responds, stopping early on a page it has not understood. A
demonstration can only show recovery from the mistake the demonstrator made, so
imitation supplies repairs for errors the student will seldom meet and none for
the ones it will. Reinforcement learning optimises the outcome we grade on the
policy's own rollouts, so the failures it learns to recover from are its own.

\subsection{Why the live web is not an RL environment}
\label{sec:rl-why}

Reinforcement learning needs an environment it can drive at scale: an exact
reset to a known state, throughput to sample in parallel, and a reward it can
trust, sustained over a run that replays the same task thousands of times. The
live web supplies none of them. Table~\ref{tab:rl-why} sets the requirements
against both candidate substrates.

\begin{table}[tbp]
\centering
\small
\caption{The requirements reinforcement learning places on an environment, and
how the two candidate substrates meet them.}
\label{tab:rl-why}
\begin{tabular}{@{}p{3.4cm}p{4.6cm}p{5.1cm}@{}}
\toprule
\textbf{Requirement} & \textbf{Live web} & \textbf{Echoverse} \\
\midrule
Exact reset to a known state
& None. Listings, dates and layouts move, so no two rollouts of the same task begin alike.
& Per-task database snapshot and restore (Sec.~\ref{sec:reset}). \\
\addlinespace[2pt]
Episode volume
& Hosts throttle and block automated traffic well before the rate RL requires.
& Self-contained applications, replicated and run in parallel. \\
\addlinespace[2pt]
Trustworthy reward
& No ground truth is exposed, so the outcome must be inferred from a screenshot by a second model.
& The grounded verifier of Sec.~\ref{sec:grading}: the outcome is compared against a reference read out of the database, not inferred from the trajectory. \\
\addlinespace[2pt]
Stationarity within a run
& Pages are redesigned and listings roll forward, so a task's meaning drifts during training.
& Fixed seed data; a task means the same thing on the first episode and the last. \\
\addlinespace[2pt]
Safety of destructive actions
& Writes hit real accounts, irreversibly.
& Safe to break and quick to reset; a corrupted world is discarded rather than repaired. \\
\bottomrule
\end{tabular}
\end{table}

\begin{figure}[tbp]
  \centering
  \includegraphics[width=\linewidth]{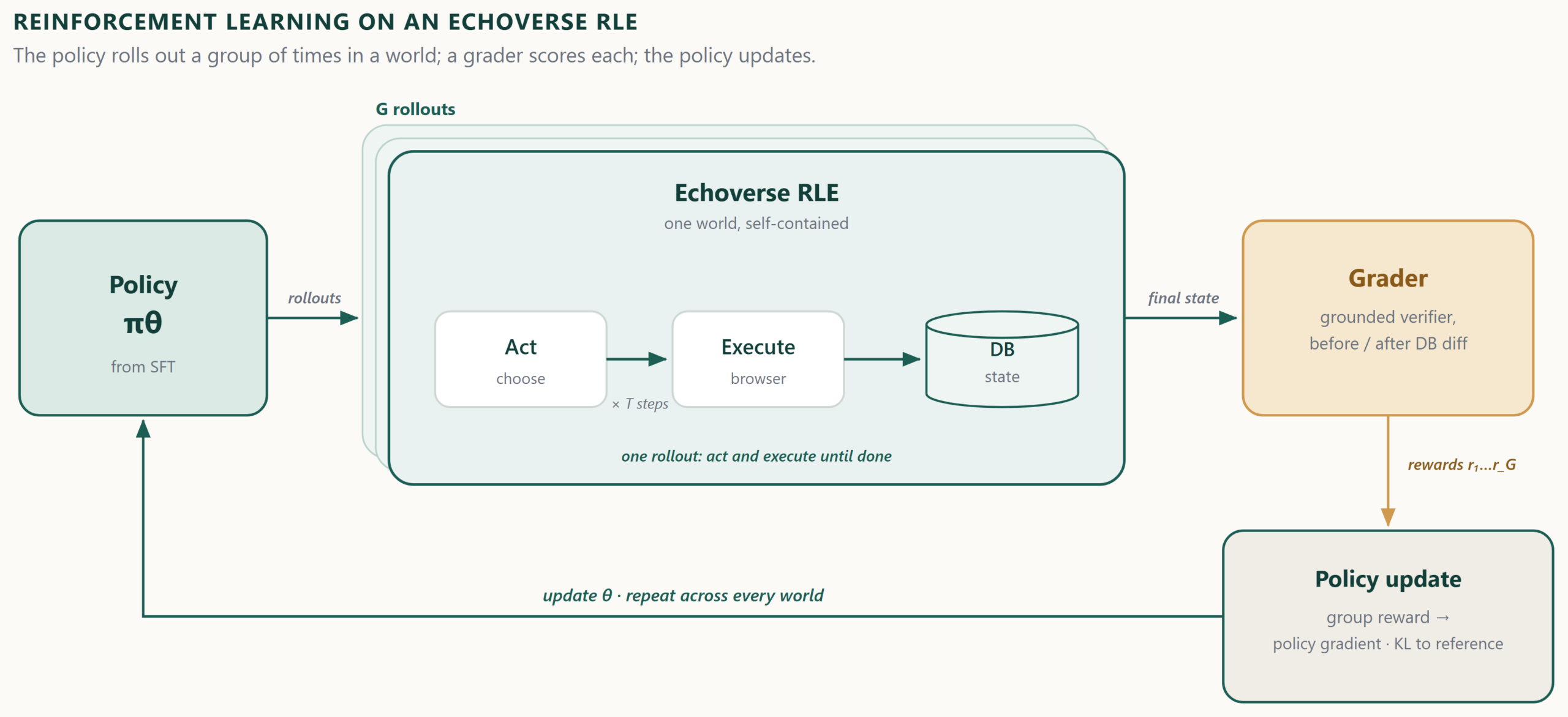}
  \caption{Reinforcement learning on an Echoverse world. From the supervised
  policy we roll out a group of trajectories in one environment; each is a
  sequence of act-and-execute steps that changes the database. A grader, the
  same grounded verifier that filtered the supervised data, sits outside the
  environment and scores each rollout's final database state into a reward. The
  group of rewards updates the policy, and the loop repeats across every
  training world. The diagram shows the grounded trajectory grader only; the run
  also applies a dense per-step reward described in Sec.~\ref{sec:rl-method}.}
  \label{fig:rlsetup}
\end{figure}

Echoverse meets all five by construction and needed no new machinery to do so:
the reset primitive is the per-task database isolation introduced in
Sec.~\ref{sec:reset} for reproducibility, and the reward is the verifier of
Sec.~\ref{sec:grading} that already filtered the supervised data. The same
worlds that benchmark an agent train one. Figure~\ref{fig:rlsetup} shows the
resulting loop.

\subsection{Method}
\label{sec:rl-method}

We take $\pisft$ as the starting policy and optimise with a group-relative
policy gradient \citep{deepseekmath2024}, producing $\pirl$, which avoids
training a value network over long multimodal browser trajectories. A
low-variance KL penalty against a fixed reference anchor keeps the policy near a
checkpoint that already speaks the action format.

\Needspace*{8\baselineskip}
Each rollout earns two rewards, summed with unit weight:
\begin{itemize}[leftmargin=1.4em,itemsep=1pt,topsep=2pt]
\item a \textbf{trajectory reward} from the database-grounded verifier of
      Sec.~\ref{sec:grading}, with GPT-4.1 making the closed comparison between
      the observed outcome and the reference read out of the database, under the
      prompts of Appendix~\ref{app:prompts};
\item a \textbf{dense per-step reward}, binary in each turn, from a multimodal
      judge (GPT-4.1 vision) that sees the pre-action screenshot with the click
      target marked. A turn scores 1 when the action demonstrably took effect,
      that is when state or address changed and the following turn builds on it,
      and 0 for a repeat, a no-op, an error, a loop, aimless scrolling, or a
      \texttt{terminate} fired before the task was done. The judge is instructed
      to compare consecutive screenshots and to disregard the agent's own
      account of what it did, which is otherwise the most common source of
      unearned credit.
\end{itemize}

The split is deliberate, and worth stating plainly because it qualifies the
grounding argument made elsewhere in this paper. The database settles whether an
episode succeeded, which is the quantity we care about, but it is largely silent
in the middle of an episode: most intermediate states are neither right nor
wrong, so a purely terminal reward gives a fifty-turn trajectory a single bit of
signal. The dense term supplies the per-step shaping the database cannot, at the
cost of reintroducing a judged signal. Three couplings keep the grounded term in
charge. The final step's dense reward is forced to zero whenever the trajectory
verdict is wrong, so process credit cannot be farmed by guessing when to stop;
terminating with the wrong outcome carries an explicit penalty; and the dense
term is applied to training rollouts only, so validation is scored on the
grounded trajectory reward alone and the held-out figures in
Sec.~\ref{sec:rl-runs} are pure outcome.

\paragraph{Credit assignment over steps.} A rollout is not one training example.
Each turn becomes its own row, tagged with its step index and with the
identifier of the prompt it came from, so a $T$-turn trajectory emits $T$ rows
and a single optimisation step yields far more rows than trajectories. The
trajectory reward is broadcast onto every row of its rollout and the step reward
is added to the row it belongs to, giving a per-row reward
\[
  R_{\mathrm{row}} \;=\; R_{\mathrm{traj}} \;+\; \lambda\,R_{\mathrm{step}},
  \qquad \lambda = 1, \quad R_{\mathrm{step}} \in \{0,1\}.
\]
The departure from the usual formulation is what the group contains. A vanilla
group-relative method groups the $G$ rollouts of a prompt and uses only their
final rows. We widen the group to every row that shares a prompt identifier,
across all $G$ rollouts and all of their turns, and centre on that group:
\[
  A_{\mathrm{row}} \;=\; R_{\mathrm{row}} - \mathrm{mean}(R_{\mathrm{group}}).
\]
We do not divide by the group standard deviation, following
\citet{drgrpo2025}: normalising by it up-weights groups that happen to be nearly
unanimous, which in this setting means the easiest and the hopeless tasks, and
it would also destroy the absolute scale on which the two reward terms are
commensurable. A single normalisation then carries both signals. A successful
trajectory lifts every row it produced, which is the outcome credit a
group-relative method already gives; and within the same group a turn scored 1
sits exactly $\lambda$ above an otherwise identical turn scored 0, so productive
turns are preferred inside losing trajectories as well as winning ones.

\subsection{Setup}
\label{sec:rl-details}

We train on five worlds, \echo{EchoBank}, \echo{EchoForge}, \echo{EchoForum},
\echo{EchoStay} and \echo{EchoTunes}, one from each workflow category of
Table~\ref{tab:suite}, on 32 GPUs.

\begin{table}[tbp]
\centering
\small
\caption{Reinforcement-learning task pool. Train rows include duplication from
upsampling; unique counts do not. Validation is a held-out set per world with
zero overlap against training.}
\label{tab:rl-data}
\begin{tabular}{@{}lrrrr@{}}
\toprule
\textbf{World} & \textbf{Train rows} & \textbf{Train unique} & \textbf{Validation} & \textbf{Overlap} \\
\midrule
\echo{EchoTunes} &  60 &  41 (upsampled) & 25 & 0 \\
\echo{EchoBank}  &  60 &  35 (upsampled) & 25 & 0 \\
\echo{EchoForum} &  70 &  70             & 25 & 0 \\
\echo{EchoStay}  &  87 &  87             & 25 & 0 \\
\echo{EchoForge} & 120 & 120             & 25 & 0 \\
\midrule
\textbf{Total}   & \textbf{397} & \textbf{353} & \textbf{125} & \textbf{0} \\
\bottomrule
\end{tabular}
\end{table}

\begin{table}[tbp]
\centering
\footnotesize
\setlength{\tabcolsep}{4pt}
\setlength{\abovecaptionskip}{4pt}
\renewcommand{\arraystretch}{0.95}
\caption{Reinforcement-learning configuration.}
\label{tab:rl-hparams}
\begin{tabular}{@{}L{4.5cm}L{8.4cm}@{}}
\toprule
\textbf{Setting} & \textbf{Value} \\
\midrule
\multicolumn{2}{@{}l}{\emph{Policy}} \\
Initialisation & $\pisft$ \\
Algorithm & Group-relative policy gradient; advantage not normalised by group standard deviation \\
Learning rate & $1\times10^{-7}$ \\
KL coefficient & $0.001$, low-variance estimator, fixed anchor \\
Entropy coefficient & $0$ \\
Off-policy correction & Token-level rollout importance sampling, threshold $2.0$ \\
\addlinespace[3pt]
\multicolumn{2}{@{}l}{\emph{Batching}} \\
Prompts per step / group size & 16 / 8 (128 trajectories per step); micro-batch 2 \\
\addlinespace[3pt]
\multicolumn{2}{@{}l}{\emph{Rollout}} \\
Temperature / top-$p$ & $0.7$ / $1.0$, top-$k$ disabled; repetition penalty $1.1$ \\
Max turns & 50 \\
Response / prompt length & 1024 / 16000 tokens, 25000 max context \\
\addlinespace[3pt]
\multicolumn{2}{@{}l}{\emph{Reward}} \\
Composition & trajectory reward $+\,1.0\times$ step reward \\
Trajectory judge & GPT-4.1 against the database reference \\
Per-step judge & GPT-4.1 vision, binary per turn, 768\,px pre-action screenshot, target marked \\
Credit assignment & one row per turn; advantage centred over all step-rows sharing a prompt \\
Couplings & final step reward zeroed on wrong outcome; wrong-termination penalty $0.1$; abort penalty on the final step only; dense term on training rollouts only \\
\addlinespace[3pt]
\multicolumn{2}{@{}l}{\emph{Schedule}} \\
Training length & a little over two epochs, 25 steps per epoch \\
Validation / checkpoint & every 5 / every 10 steps; reported checkpoint at the two-epoch mark, step 50 \\
\bottomrule
\end{tabular}
\end{table}

\paragraph{Task pool.} The pool is filtered, balanced and disjoint from
evaluation. We sample the starting policy four times on every candidate task and
keep only those it solves one, two or three times out of four. Both extremes are
discarded: a task solved zero times out of four gives a group of rollouts that
is uniformly wrong, and a task solved four times out of four gives one that is
uniformly right, and in either case the advantage in Sec.~\ref{sec:rl-method} is
identically zero and the task contributes no gradient. Retaining only the
mid-range does not guarantee a mixed group, since pass@4 records the behaviour
of the starting policy rather than a property of the policy at the step where
the task is drawn, but it makes one substantially more likely. Within that pool
we balance per-world counts into a band of 60 to 120 rows: \echo{EchoTunes} and
\echo{EchoBank} fall below the floor and are upsampled by duplication,
\echo{EchoForge} is capped from 237 to 120 by fixed-seed subsampling.
Table~\ref{tab:rl-data} gives the resulting mix. The filter does not make the
pool easy: of the 353 unique tasks, 189 are labelled hard, 124
medium and 40 easy, and 288 of them require a write
(180 \textsc{read\_write}, 108 \textsc{write}, 65 \textsc{read}). The largest
categories are multi-step planning (112 tasks), messaging (38), playlists (25),
wishlists (21) and community actions (20).

\paragraph{Optimisation and rollouts.} Table~\ref{tab:rl-hparams} lists the
configuration. A step draws 16 prompts at group size 8, giving 128 trajectories
per step collected in a single wave. Rollouts are sampled at temperature 0.7
with a mild repetition penalty and capped at 50 turns. Note that this is half
the 100-action budget used for every evaluation in Sec.~\ref{sec:experiments}
and for the held-out validation reported below, so the policy is trained under a
tighter budget than it is scored under. Validation is greedy, so the reported
score is not inflated by sampling.

\FloatBarrier

\subsection{Results}
\label{sec:rl-runs}

\begin{figure}[!ht]
  \centering
  \setlength{\abovecaptionskip}{4pt}
  \includegraphics[width=\linewidth]{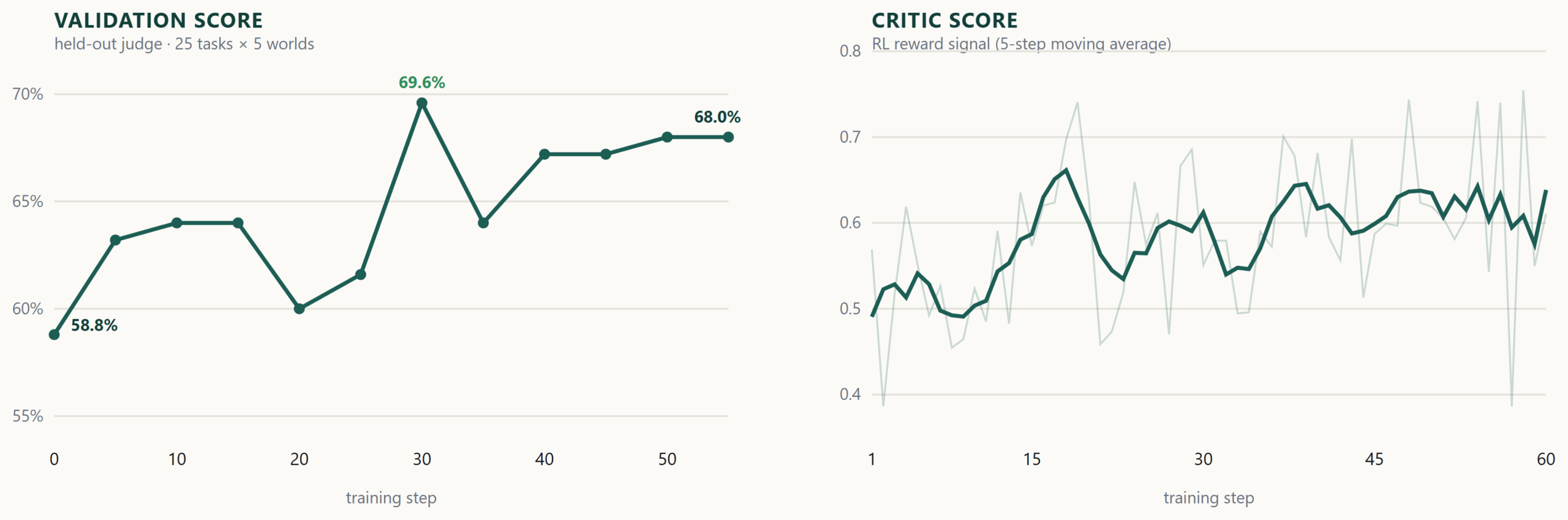}
  \caption{Reinforcement learning on five worlds, a little over two epochs.
  Left: the held-out judged score, 25 tasks per world graded by the
  database-grounded verifier. Right: the training reward, which unlike the
  judged score includes the dense per-step term.}
  \label{fig:rlcurves}
\end{figure}

We report two quantities. The \emph{judged score} is the mean trajectory reward
over the held-out set, the grounded verifier's verdict of
Sec.~\ref{sec:grading}, scored on Table~\ref{tab:rl-data}'s validation column.
The \emph{training reward} is the mean total reward earned on training rollouts,
and so includes the dense per-step term. The judged score rises from $58.8\%$ to
$68.0\%$ at the two-epoch mark, step 50, and the training reward trends upward
over the same period (Figure~\ref{fig:rlcurves}).

\FloatBarrier

\section{Conclusion}
\label{sec:release}

A world is more than a benchmark to score against: it is something the loop
keeps improving, because the same graded run that measures the model also shows
what to fix in the world. Deep worlds transfer where shallow clones pull
capability down; one control rebuilt in a hundred forms teaches a skill that
carries to held-out widgets and to the open web; co-evolution moves model and
world together; and reinforcement learning against the same worlds takes the
policy from $58.8\%$ to $68.0\%$ on held-out tasks.

What matters is not the largest inventory of synthetic websites but a factory
that finds what an agent cannot yet do, builds or repairs the world that teaches
it, and runs the loop again. More deep worlds for the closed domains public
benchmarks cannot reach, more capability worlds for the interactions models keep
failing, and above all more reinforcement learning against grounded worlds:
these interact, since broader worlds admit longer training runs and every round
exposes the next capability to build.

We release environment code and graded test tasks for four worlds: the deep
domains \echo{EchoStay} and \echo{EchoForge}, and the two capability worlds with
their in-distribution and held-out splits. Every task carries the
database-grounded verifier that scores it, so an agent is measured against
application state rather than against a judge, and we release evaluation tasks
only so they stay uncontaminated as a measure.
Code: \url{https://aka.ms/echoverse}.

\begin{ack}
We thank Alexey Taymanov, Andrew Zhao, Aravind Rajeswaran, Luiz Do Valle,
Pashmina Cameron, Rafah Hosn, Spencer Whitehead, Zach Nussbaum and Yadong Lu for
their valuable help, insightful discussions, and continued support throughout
this work.
\end{ack}

\clearpage
\bibliographystyle{plainnat}

\bibliography{echoverse}

\appendix

\section{Naming conventions}
\label{app:naming}

Every world in this paper is built on a behaviourally faithful clone of a class
of application rather than of any particular product, and is named for the
workflow it represents. Public benchmarks and public data sources are referred
to by their real names, since they are cited rather than cloned.

\section{The supervised corpus}
\label{app:corpus}

Table~\ref{tab:corpus} gives the composition of the supervised fine-tuning
corpus of Sec.~\ref{sec:sft}, the single body of data behind every $\pisft$
result in Sec.~\ref{sec:experiments}. Every trajectory in it is a GPT-5.4
rollout that the grounded verifier of Sec.~\ref{sec:grading} scored as passing;
rollouts that failed are discarded rather than kept as negatives. The per-world
counts are uneven because the worlds are, in how many distinct task sets they
support and in how long a task takes: an \echo{EchoStay} booking runs to nearly
34 actions on average while a nested-filter submission takes fewer than five.
The trajectory-scaling sweep of Sec.~\ref{sec:exp-scaling} subsamples this same
corpus and holds these proportions fixed at every point, so the only quantity
that varies along that axis is volume.

\begin{table}[!ht]
\centering
\small
\caption{The supervised corpus, per world. \emph{Task sets} counts the
separately exported task datasets from that world that contribute to the corpus.
Steps are agent actions; a trajectory ends when the agent emits
\texttt{terminate} or exhausts its budget.}
\label{tab:corpus}
\begin{tabular}{@{}lrrrr@{}}
\toprule
\textbf{World} & \textbf{Task sets} & \textbf{Trajectories} & \textbf{Steps} & \textbf{Avg.\ steps/traj.} \\
\midrule
\echo{EchoMail}      & 5 & 4{,}482 & 91{,}155 & 20.34 \\
\echo{EchoCalendar}  & 3 & 4{,}109 & 91{,}891 & 22.36 \\
\echo{EchoChat}      & 1 &   832 & 14{,}954 & 17.97 \\
\echo{EchoML}        & 4 & 2{,}450 & 70{,}397 & 28.73 \\
\echo{EchoForge}     & 1 &   824 & 20{,}596 & 25.00 \\
\echo{EchoBank}      & 2 &   852 &  7{,}801 &  9.16 \\
\echo{EchoCare}      & 1 &   703 & 13{,}680 & 19.46 \\
\echo{EchoForum}     & 3 & 1{,}472 & 25{,}149 & 17.08 \\
\echo{EchoTunes}     & 2 &   919 & 15{,}759 & 17.15 \\
\echo{EchoStay}      & 3 &   757 & 25{,}586 & 33.80 \\
Datepicker world     & 3 & 1{,}827 & 19{,}475 & 10.66 \\
Nested-filter world  & 3 & 1{,}782 &  8{,}371 &  4.70 \\
\midrule
\textbf{Total}       & \textbf{31} & \textbf{21{,}009} & \textbf{404{,}814} & \textbf{19.27} \\
\bottomrule
\end{tabular}
\end{table}

\FloatBarrier

\section{Verifier prompts}
\label{app:prompts}

The comparison described in Sec.~\ref{sec:grading} is a single LLM call per
task, made with the prompts reproduced below. Neither judge receives the
trajectory, the screenshots, or the agent's own account of what it did. The read
judge sees the goal, the reference answer and the agent's answer. The write
judge sees the goal, the reference state change and the \texttt{sqldiff} between
the pristine per-task database $D_0$ and the final database $D_T$, which is what
makes the write check a statement about the change rather than about the end
state alone. Both return a binary score with a short justification.

\subsection{Read tasks}

\begin{lstlisting}[style=prompt]
You are an evaluation judge. You compare an agent's answer to a reference answer for a Website/application task.

The task asks the agent to look up information (a "read" task). You must decide
whether the agent's answer is semantically equivalent to the reference answer.

Rules:
- Minor formatting differences (extra whitespace, different date formats, quoting) are OK.
- Synonyms and paraphrases are OK as long as the meaning is the same.
- Monetary amounts may be rounded (e.g., $287.62 and $288 are equivalent; $181.63 and $182 are equivalent). Small rounding differences in prices, fees, or totals are acceptable because the agent reads displayed values from the UI which may round differently than the database.
- Numeric values may be expressed in equivalent units or human-readable formats. For example: 157286400 bytes and "150 MB" are equivalent; 2500000000 and "2.5 billion" or "2,500,000,000" are equivalent; 42000000 and "42M" or "42,000,000" are equivalent. Accept any representation that conveys approximately the same quantity (within reasonable rounding for unit conversion).
- Counts and IDs should match, but minor differences due to UI display rounding or pagination are acceptable if the values are close.
- Email addresses and names must match (case-insensitive).
- If the agent provides MORE information than needed but includes the correct answer, that counts as correct.
- If the agent includes additional context (e.g., a date alongside a time) but the core answer is correct, that counts as correct.
- If the agent's answer is completely wrong or missing the key information, it fails.

Return a JSON object with exactly these keys:
{
  "score": <int 0 or 1>,
  "reasoning": "<brief explanation>"
}

score guide:
  1 = answer is semantically correct
  0  = no required changes present or completely wrong changes
\end{lstlisting}

\noindent The user message is instantiated per task as:

\begin{lstlisting}[style=prompt]
## Task goal
$question

## Reference answer (ground truth)
$ref_answer

## Agent's answer
$agent_answer

Compare the agent's answer against the reference answer and return your JSON verdict.
\end{lstlisting}

\subsection{Write tasks}

\begin{lstlisting}[style=prompt]
You are an evaluation judge for a Website/application. You determine whether
a task was completed correctly by comparing the intended state change with
the actual database changes.

You receive:
1. The task goal (what the agent was supposed to do)
2. The expected state change (reference_state_change)
3. The actual SQL diff (output of sqldiff showing what changed in the database)

The SQL diff contains SQL statements that would transform the BEFORE database
into the AFTER database. Common patterns:
- UPDATE table SET col=val WHERE ...  -> a field was changed
- INSERT INTO table ...               -> a new row was added
- DELETE FROM table WHERE ...          -> a row was removed

Rules:
- The agent may have made additional changes beyond what was required (e.g.,
  marking an email as read while also starring it). Extra changes are OK as
  long as the required change was made.
- If the diff is empty, the agent made no changes -> likely a failure.
- Focus on whether the REQUIRED changes (from reference_state_change) are
  present in the diff.

Return a JSON object with exactly these keys:
{
  "score": <int 0 or 1>,
  "reasoning": "<brief explanation of what matched and what didn't>"
}

score guide:
  1  = all required changes are present
  0  = no required changes present or completely wrong changes
\end{lstlisting}

\noindent The user message is instantiated per task as:

\begin{lstlisting}[style=prompt]
## Task goal
$goal

## Expected state change (reference)
$ref_state_change

## Actual database diff (sqldiff output)
$sql_diff

Evaluate whether the required changes were made. Return your JSON verdict.
\end{lstlisting}

\section{Capability-world widget catalogue}
\label{app:widgets}

Every widget family the two capability worlds render, with the interaction each
one demands of the agent. Table~\ref{tab:widgets} covers the nested-filter world
and Table~\ref{tab:widgets-date} the datepicker world. Held-out families never
appear in training and are used only for evaluation.

\begin{table}[htbp]
\centering
\small
\setlength{\tabcolsep}{4pt}
\caption{Nested-filter widget families. Training covers 20 families over 200
generated frontends in 15 visual styles; the held-out group covers 9 compound
families over 100 frontends in 9 styles.}
\label{tab:widgets}
\begin{tabular}{@{}L{2.5cm}L{4.05cm}L{2.55cm}L{3.9cm}@{}}
\toprule
\multicolumn{2}{@{}l}{\textbf{Training (in-distribution)}} & \multicolumn{2}{l}{\textbf{Held out}} \\
\cmidrule(r){1-2}\cmidrule(l){3-4}
\textbf{Family} & \textbf{Interaction} & \textbf{Family} & \textbf{Interaction} \\
\midrule
\addlinespace[1pt]
Calculator and menu & Compute a value, pick from a dropdown & Booking search & Destination with check-in and check-out dates \\
Calendar filter & Pick a date to filter by & Override calculator & Calculator with default overrides \\
Filter dialog & Open an ``all filters'' modal & Calendar combination & Calendar plus secondary controls \\
Search field & Type a free-text query & Cost calculator & Region and usage cost estimator \\
Facet checkboxes & Tick multiple category boxes & Property facets & Beds, baths and price facets \\
Range slider & Set a minimum and maximum & Modal filter flow & Multi-step modal filters \\
On/off toggles & Boolean switches & Strict facets & Exact-match facet filter \\
Filter chips & Click pill toggles on and off & Adoption filters & Species, compatibility and sex \\
Calculator field & Type or derive a numeric threshold & Full property panel & Complete real-estate filter panel \\
Active-filter bar & Chips of the applied filters & & \\
\bottomrule
\end{tabular}
\end{table}

\begin{table}[htbp]
\centering
\small
\setlength{\tabcolsep}{4pt}
\caption{Datepicker widget families. Training covers 6 core types over 10
contexts, 100 generated frontends in 6 visual styles; the held-out group covers
10 unusual widgets over 36 scenarios, 80 frontends in 9 styles.}
\label{tab:widgets-date}
\begin{tabular}{@{}L{2.5cm}L{4.05cm}L{2.55cm}L{3.9cm}@{}}
\toprule
\multicolumn{2}{@{}l}{\textbf{Training (in-distribution)}} & \multicolumn{2}{l}{\textbf{Held out}} \\
\cmidrule(r){1-2}\cmidrule(l){3-4}
\textbf{Family} & \textbf{Interaction} & \textbf{Family} & \textbf{Interaction} \\
\midrule
\addlinespace[1pt]
Date and time & A date plus a time of day & Calendar heatmap & Grid coloured by density \\
Single date & Pick one calendar date & Scroll wheel & Spinning wheel columns \\
Date range & A start and end span & ISO week picker & Select a whole calendar week \\
Bounded date & Date within minimum and maximum limits & Multi-date & Pick several dates at once \\
Date of birth & Birthdate entry & Duration stepper & Step an ISO duration up or down \\
Month and year & Month and year only & Time-only clock & Hours, minutes and seconds, no date \\
\addlinespace[2pt]
\multicolumn{2}{@{}l}{\emph{(no further training families)}} & Quarter picker & Select a fiscal quarter \\
 & & Fiscal-year picker & Pick a fiscal year by decade \\
 & & Timeline slider & Slide along a date axis \\
\bottomrule
\end{tabular}
\end{table}

\end{document}